%% file: main.tex
\pdfoutput=1
\documentclass[10pt]{article}
\usepackage[margin=0.88in]{geometry}
\usepackage[T1]{fontenc}
\usepackage{mathpazo,microtype,amsmath,amssymb,graphicx,booktabs,array,longtable}
\usepackage[font=small,labelfont=bf]{caption}
\usepackage{xcolor,placeins,enumitem}
\usepackage[colorlinks=true,allcolors=blue!55!black]{hyperref}
\usepackage[numbers,sort&compress]{natbib}
\hypersetup{pdftitle={Do Multimodal Models Decide Consistently on Equivalent Optical Spectra?},pdfauthor={Ziyuan Wang, Genliang Zhu, Xiaobo Chen},pdfkeywords={multimodal language models, decision models, optical spectra, scientific data, tables and plots, representation invariance, photon energy, abstention, unanswerable questions, evidence sufficiency, LoRA fine-tuning, shortcut learning, AI for science}}
\newcommand{\scope}{D_{\mathrm{scope}}}
\newcommand{\outside}{\Delta_{\mathrm{out}}}
\makeatletter
\let\inputrows\@@input
\makeatother
\title{Do Multimodal Models Decide Consistently\\on Equivalent Optical Spectra?}
\author{%
Ziyuan Wang\textsuperscript{1}, Genliang Zhu\textsuperscript{2}, Xiaobo Chen\textsuperscript{3}\\[0.5em]
\small \textsuperscript{1}Independent Researcher\\
\small \textsuperscript{2}Georgia Institute of Technology\\
\small \textsuperscript{3}University of Massachusetts Amherst\\[0.4em]
\footnotesize \textsuperscript{1}\href{mailto:ziyuan.wang.engineer@gmail.com}{\nolinkurl{ziyuan.wang.engineer@gmail.com}}\\
\footnotesize \textsuperscript{2}\href{mailto:sebby@gatech.edu}{\nolinkurl{sebby@gatech.edu}}\\
\footnotesize \textsuperscript{3}\href{mailto:xchen@umass.edu}{\nolinkurl{xchen@umass.edu}}}
\date{}

\begin{document}
\maketitle
\vspace{-3.2em}
\begin{abstract}
A spectrum can be given to a model as a table in nanometers or as a plot in photon energy, and the decision should not depend on which one it sees. When the data needed for an answer are missing, the model should say so. We test representation invariance and abstention on measured optical spectra of silicon metasurfaces (MOCLIP) and gold nanoparticles, with labels computed from the data. Each spectrum is shown as four tables (nm, $\mu$m, scientific notation, eV) and two plots (nm, eV), and partial spectra remove data inside or outside the queried range. We evaluate the decision models d3-nano and Jev~1.13, three GPT-6 Luna configurations and the Qwen3.5-2B base model. Changing only the representation moves accuracy by up to 41 percentage points within one model, mostly on photon-energy tables and plots. With the same evidence, asking in photon energy instead of wavelength helps some model--view pairs by 16--50 points and hurts others by up to 25. When the relevant extremum is removed, unadapted models rarely answer ``undetermined''. LoRA fine-tuning of d3-nano improves comparisons on the trained table formats, but much of its gain on unanswerable questions comes from a shortcut: it abstains whenever the input is truncated. Training with answerable truncations avoids the shortcut in all six seeds, with a 90--100 point gap between abstention after inside and outside cuts on held-out structures; with the same amount of extra original-task data, five of six seeds keep the shortcut. On gold spectra the two training sets perform similarly. Multimodal language models used for scientific decisions should be tested on equivalent representations and on missing evidence, not only on average accuracy. We release the tasks, per-response results and analysis code.
\end{abstract}

\section{Introduction}
An optical spectrum can be given as a numerical table or a plot. Its horizontal axis can show wavelength in nanometers or micrometers, or photon energy. The physical response is the same, but reading and interpreting these forms requires different steps. For example, a red shift means a longer wavelength and a lower photon energy. Also, the lowest point visible in a partial spectrum may not be the lowest point in the range asked about. A useful decision model should keep its answer under equivalent representations, change it when the question requires a different answer, and abstain when necessary evidence is missing.

Native decision models receive a set of alternatives and return a choice with probabilities \citep{almeida2026jev,openai2026decisions,d3nano2026}. This format is convenient for automated scientific workflows. However, a valid output format and high average accuracy do not establish whether the model follows the physical meaning of the data or recognizes that part of the required evidence is missing.

We test both issues using measured optical spectra and labels calculated directly from the arrays. For each original task, we vary the number format, coordinate, modality, evidence order, question direction, or data coverage. Each change has a defined expected answer. To test evidence sufficiency, we compare a complete spectrum with two partial views of the same spectrum. Removing data outside the queried interval leaves the answer available. Removing the segment containing the relevant extremum makes the question unanswerable. We then use low-rank adaptation (LoRA) \citep{hu2021lora} to examine which errors an open decision model can learn to correct. The study has three parts:
\begin{itemize}[leftmargin=1.3em,itemsep=1pt,topsep=2pt]
\item We compare tables in wavelength, micrometers, scientific notation, and photon energy, together with wavelength and energy plots. All use the same underlying arrays. We reverse the question direction to check cases where the answer should change, and vary wording while keeping the energy-coordinate evidence fixed.
\item We remove data inside or outside the queried range of the same spectrum. This tests whether abstention depends on which evidence is missing, rather than simply on whether the input is truncated.
\item We vary the representations, example count, training duration, and answerable truncations used for training. We report all seeds and include controls with a similar amount of added data and a base model with the same backbone.
\end{itemize}
The largest representation effects occur in energy coordinates. Energy wording helps some model--task combinations and hurts others. Unadapted models seldom abstain when relevant evidence is missing, and most adaptation gains are limited to the representations and dataset used for training.

\section{Related work}
\paragraph{Optical response prediction and inverse design.}
Neural models have also been used to predict spectra from nanostructure geometry. \citet{lin2025transfer} transferred a model trained on annular aperture arrays to nanohole arrays to improve prediction with limited target data. \citet{wang2025gold} used a single network to predict transmission spectra for gold nanodisks, truncated cones, and nanocones. \citet{chen2025annular} combined transmission prediction for annular aperture arrays with a geometry search for inverse design. These studies learn mappings between structure and optical response. Our study tests decisions made from measured spectra, including whether the answer changes with the representation or when relevant evidence is missing.

\paragraph{Equivalent representations and modality gaps.}
IsoBench gives the same problems as images, text, and formal notation. It finds that multimodal models generally perform better with text \citep{fu2024isobench}. MaCBench reports a similar result for chemistry and materials data, including a large text advantage in reading X-ray diffraction peak positions \citep{alampara2025macbench}. For some table-reasoning tasks, however, table images match or outperform text formats \citep{deng2024tables}. Chart robustness studies keep the data fixed and change scale, colour, or layout \citep{mukhopadhyay2024unraveling}. We examine coordinate changes that also require a physical direction relation, such as longer wavelength and lower photon energy. We include controls where the answer should change and where only the question wording changes.

\paragraph{Chart and spectral question answering.}
ChartQA and ChartQAPro test visual and logical reasoning about charts, with unanswerable questions included in the latter \citep{masry2022chartqa,masry2025chartqapro}. SpecVQA uses expert questions about seven types of scientific spectra \citep{shen2026specvqa}. We use fixed templates and spectra with clear extrema. This makes the comparisons easier to control, but covers a narrower range of questions than practical use.

\paragraph{Evidence sufficiency and abstention.}
Whether evidence is sufficient depends on the question. Retrieval-augmented models often answer despite insufficient context. Fine-tuning on sufficiency labels can increase abstention but also reduce the number of correct answers \citep{joren2025sufficient}. Easily constructed unanswerable examples can teach cues that fail on harder cases \citep{abdumalikov2024answerability}, a form of shortcut learning \citep{geirhos2020shortcut}. TrustDABench tests refusal when table changes remove necessary evidence, together with invariance under changes that preserve the answer, including unit conversion \citep{shi2026trustdabench}. In our answerable control, we delete an irrelevant part of the same spectrum. We then test whether adding these examples to training changes abstention behavior. This use of controlled input changes follows the approach of behavioral testing frameworks such as CheckList \citep{ribeiro2020checklist}.

\paragraph{Decision interfaces and option bias.}
Multiple-choice models can prefer particular option identifiers \citep{zheng2024selection}. In typed decision models, option labels can outweigh the definitions supplied by the caller \citep{azizi2026labels}. Effects of thresholds and option order also vary across systems \citep{pai2026jevbench}, and gains from language-side adaptation may not transfer to other settings \citep{zhou2026pixeljev}. Other work evaluates scientific semantic decisions through their downstream effects \citep{deng2026jev}. Here we test evidence order, coordinate meaning, and the queried range. We also compare one decision model with its base model.

\section{Data and paired task construction}
\subsection{Measured optical data and deterministic labels}
MOCLIP contains transmission spectra from 10,000 fabricated silicon-on-glass metasurfaces measured under two polarizations \citep{rodionov2026moclippaper,rodionov2026moclip}. Each spectrum has 29 samples from 450 to 730\,nm, spaced by 10\,nm. The main evaluation uses $y$ polarization. We define the target dip as the global minimum on this grid and do not fit a sub-grid minimum. We keep interior dips with prominence of at least 0.15. Their transmission must differ by at least 0.05 from the next local dip with prominence of at least 0.05. Band questions also require a 15\,nm margin from internal band boundaries. These filters give clear labels and exclude overlapping or unresolved resonances.

The second dataset contains 448 peptide-assisted gold-nanoparticle syntheses with measured UV--visible absorbance \citep{lachowski2022multivariate,lachowski2022gold}. Wavelengths run from 300--995\,nm in 5\,nm steps. We use background-subtracted absorbance over 450--800\,nm, and a clean-peak filter keeps 131 spectra. Gold peaks are often broad, so the questions ask which side of a threshold contains the maximum. The main threshold is 542.5\,nm. We require at least 10\,nm separation from it and a 3\% peak-absorbance margin over the other side. Comparison questions require a peak shift of at least 25\,nm.

\subsection{Task instances, variants, and representations}
An \emph{original instance} uses one spectrum, a pair, or three candidates. A \emph{variant} changes one part of its question or evidence, and a \emph{request} presents one variant in one representation (Table~\ref{tab:data}). MOCLIP has four task types, each with 48 original instances. \emph{Extreme} asks which band contains the minimum or maximum of the same spectrum. \emph{Compare} asks which of two dips lies at longer or shorter wavelength, with a minimum separation of 40\,nm. We use both evidence orders and both question directions. \emph{Specification} asks which of three candidates meets a dip-location window and a depth threshold. One wrong candidate is deep enough but outside the window; the other is inside but too shallow. The correct candidate rotates through A, B, and C. \emph{Sufficiency} pairs a full spectrum with a partial view that removes its true dip. Long-side, short-side, and middle deletions occur equally often. Gold has 20 threshold, 16 compare, and 18 sufficiency instances. Only sufficiency questions offer ``cannot be determined'' as an option.

\begin{table}[t]
\centering\small
\caption{Evaluation sets. Core variants exist in all six representations; energy-worded probes exist only as eV tables and eV plots. Request counts are for a model that supports images.}
\label{tab:data}
\begin{tabular}{lrrrrr}
\toprule
Set & Instances & Unique spectra & Variants & Core variants & Requests \\
\midrule
MOCLIP main & 192 & 336 & 720 & 528 & 3,552 \\
Gold main & 54 & 70 & 244 & 140 & 1,048 \\
MOCLIP scope & 54 & 54 & 162 & 162 & 972 \\
Gold scope & 60 & 60 & 180 & 180 & 1,080 \\
\bottomrule
\end{tabular}
\end{table}

We render each core variant as four tables (decimal nm, decimal $\mu$m, scientific notation in nm, and eV) and two plots (nm and eV). The conversion is $E\,[\mathrm{eV}]=1239.841984/\lambda\,[\mathrm{nm}]$. Transmission and absorbance are response ratios, so changing the horizontal coordinate does not require a spectral-density Jacobian. Answer keys are fixed, while the displayed ranges and units follow the representation. Rows in eV tables are sorted by increasing energy, which reverses the wavelength order. Their spacing is nonuniform: 0.060\,eV between 450 and 460\,nm, and 0.024\,eV between 720 and 730\,nm. Tables use rounded values. Plots are $800\times500$ raster images with labeled axes, and middle deletions appear as disconnected segments. The underlying samples and answers are equivalent, but their visual and numerical readability can differ. A deterministic program that parses the rendered tables reaches 100\% accuracy on every table request. Appendix~\ref{app:prompts} gives the original requests.

Compare questions normally ask about longer or shorter \emph{wavelength}, even when the evidence uses energy coordinates. Additional variants on eV tables and plots ask about lower or higher \emph{photon energy}. We use the same control for gold threshold questions. The evidence stays fixed, so this comparison tests the extra step of relating wavelength wording to an energy axis.

\subsection{Query-relative evidence controls}
\label{sec:scope}
\begin{figure}[t]
\centering
\includegraphics[width=.97\linewidth]{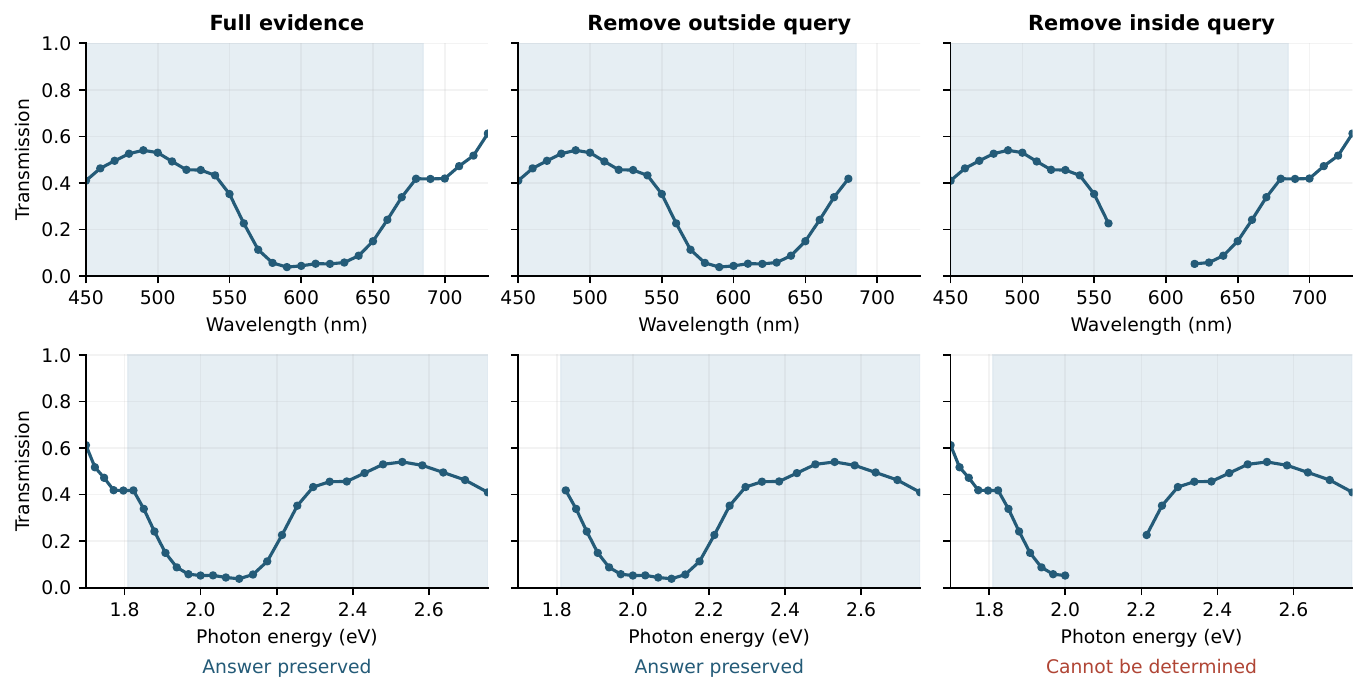}
\caption{One MOCLIP scope instance in wavelength and energy coordinates. Shading marks the queried interval and is added for illustration only; model inputs state the interval in the question.}
\label{fig:scope}
\end{figure}

Every truncated view in the main sufficiency tasks is unanswerable. A model can therefore succeed by treating any missing data as a reason to abstain. The scope sets test this possibility. Each question states an interval $Q$, and each spectrum is shown in three forms: full; with samples \emph{outside} $Q$ removed, leaving the answer available; and with the segment \emph{inside} $Q$ containing the relevant extremum removed, labeled ``cannot be determined'' (Figure~\ref{fig:scope}). MOCLIP uses query intervals of 450--685 or 605--730\,nm. Gold uses 450--650 or 490--800\,nm, with five thresholds from 522.5 to 562.5\,nm. Both sets balance long-side, short-side, and middle inside cuts, with 18 MOCLIP and 20 gold instances per type. The MOCLIP scope set uses $x$ polarization and excludes all structures used in main testing and training. It reuses some pilot structures because the clean-dip filters limit the available pool. Gold scope and main sets share 30 spectra. They are two task constructions from one dataset, not independent replications.

\section{Models and analysis}
\subsection{Model configurations}
\textbf{d3-nano} is a 2.21B-parameter multimodal decision model fine-tuned from Qwen3.5-2B. We use checkpoint \texttt{80820624a817} (v3.1.0 weights) in bfloat16 and its local probability readout \citep{d3nano2026}. \textbf{d3-nano-pa} averages the runtime outputs for forward and reversed option orders. \textbf{Jev 1.13} is a text-only decision model accessed through OpenRouter \citep{almeida2026jev}.

We test three \textbf{GPT-6 Luna} configurations through OpenRouter, with OpenAI as the only provider. These are the decision route (\texttt{openai/gpt-6-luna-decisions}; \emph{Luna Decisions}) and the generative route through Chat Completions with a strict JSON schema at reasoning effort \texttt{none} or \texttt{low} (\emph{Luna gen.\ none}, \emph{Luna gen.\ low}). The routes share a model family and provider. We cannot verify that they share checkpoints or inference implementations, so we compare the available configurations. We did not test the native OpenAI \texttt{/v1/decisions} endpoint \citep{openai2026decisions}.

As a reference with the same backbone, we test the \textbf{Qwen3.5-2B base} model \citep{qwen2026qwen35} with thinking disabled. After giving lettered options, we restrict its next-token distribution to those letters (Appendix~\ref{app:prompts}). API calls were made on 14--16 and 24 October 2026. Local models ran on RunPod GPUs.

\subsection{Metrics and uncertainty}
We calculate accuracy using the canonical answer keys. Unless stated otherwise, summaries of representation, training, and interface effects use core variants. Energy-worded probes are used only in the wording analysis. We measure each representation effect as its paired accuracy difference from the nm table on the same variants. For a representation set $R$, agreement on variant $v$ is $C_R(v)=\mathbb{1}\{\hat y_{v,r}=\hat y_{v,r'}\ \forall r,r'\in R\}$. Higher accuracy also tends to produce higher agreement. We therefore separate agreement into cases where all answers are correct and cases where all are the same but wrong. We compare the all-correct share with its expected value under independent errors across representations, using each representation's accuracy within each task type.

For the scope sets, with $U_{i,c,r}$ indicating an ``undetermined'' prediction and $A_{i,c,r}$ correctness for instance $i$, cut $c$ and representation $r$,
\begin{align}
\scope &= 100\,\mathbb{E}_{i,r}[U_{i,\mathrm{in},r}-U_{i,\mathrm{out},r}], &
\outside &= 100\,\mathbb{E}_{i,r}[A_{i,\mathrm{out},r}-A_{i,\mathrm{full},r}].
\end{align}
The ideal values are $+100$ and $0$. A model that always abstains or never abstains has $\scope=0$. Both measures are relative, so we also report full-view accuracy. The high-confidence error fraction is the proportion of all requests with $\max_k p(k)\ge0.9$ and an incorrect choice. Generative probabilities are reported by the model in JSON and should not be treated as equivalent to decision-interface probabilities.

We calculate 95\% percentile intervals from 2,000 bootstrap resamples of original task instances. Each instance's variants and representations stay together, and differences are paired within instances. We report LoRA seeds separately. The study developed through sequential pilots, so the intervals are descriptive and have no correction for multiple comparisons.

\section{Results}
\subsection{Representations change decisions}
Table~\ref{tab:accuracy} reports core accuracy for tables and plots. Table~\ref{tab:repeff} gives the paired difference between each representation and the nm table. Within wavelength coordinates, number-format changes have small effects for most models, usually within $\pm5$ points. The largest is $-12$ for d3-nano on gold scientific-notation tables. Energy coordinates cause larger changes. On gold, eV tables lower accuracy by 41 points for Jev, 21 for Luna gen.\ none, 20 for d3-nano, and 13 for Luna Decisions. On MOCLIP, low-reasoning Luna loses 10 points on eV tables despite near-ceiling accuracy in the other representations.

Plots do not affect all models in the same way. Luna gen.\ none loses 10 and 20 points on MOCLIP nm and eV plots, and 34 points on gold eV plots. This agrees with earlier findings of better performance on text \citep{fu2024isobench,alampara2025macbench}. Luna Decisions instead gains 15 and 9 points on MOCLIP and gold nm plots. For d3-nano, the nm table is the least accurate MOCLIP representation, and every other representation gives a gain of 4--7 points.

\begin{table}[t]
\centering\small
\caption{Core-variant accuracy (\%, 95\% instance-bootstrap intervals). MOCLIP has 528 core variants and gold 140. Jev has no image input. The Qwen3.5-2B base row comes from the supplementary batch.}
\label{tab:accuracy}
\begin{tabular}{lrrrr}
\toprule
& \multicolumn{2}{c}{MOCLIP} & \multicolumn{2}{c}{Gold} \\
\cmidrule(lr){2-3}\cmidrule(lr){4-5}
Model & Tables & Plots & Tables & Plots \\
\midrule
d3-nano & 56 [54, 59] & 60 [57, 63] & 64 [59, 69] & 55 [48, 63] \\
Jev & 78 [75, 81] & -- & 81 [77, 85] & -- \\
Luna Decisions & 65 [62, 67] & 70 [67, 73] & 82 [77, 87] & 80 [75, 85] \\
Luna gen. none & 74 [70, 77] & 60 [56, 63] & 86 [82, 90] & 74 [68, 79] \\
Luna gen. low & 95 [94, 96] & 93 [91, 94] & 97 [96, 99] & 96 [93, 99] \\
Qwen3.5-2B base & 40 [38, 42] & 46 [44, 49] & 44 [36, 51] & 58 [50, 65] \\
\bottomrule
\end{tabular}
\end{table}

\begin{table}[t]
\centering\small
\caption{Paired accuracy difference from the nm table on the same core variants (percentage points, 95\% intervals).}
\label{tab:repeff}
\setlength{\tabcolsep}{4pt}
\begin{tabular}{llrrrrr}
\toprule
& Model & $\mu$m table & sci.\ table & eV table & nm plot & eV plot \\
\midrule
MOCLIP & d3-nano & +4 [+2, +7] & +5 [+2, +9] & +5 [+1, +9] & +7 [+3, +11] & +7 [+3, +11] \\
 & Jev & +1 [-1, +3] & +2 [-0, +5] & -5 [-8, -1] & -- & -- \\
 & Luna Decisions & +5 [+2, +9] & +2 [-2, +6] & +0 [-4, +5] & +15 [+11, +19] & -0 [-5, +4] \\
 & Luna gen. none & +1 [-2, +4] & +1 [-2, +4] & -5 [-9, +0] & -10 [-15, -4] & -20 [-25, -13] \\
 & Luna gen. low & -2 [-3, -0] & -2 [-3, -0] & -10 [-13, -7] & -5 [-7, -3] & -6 [-9, -4] \\
 & Qwen3.5-2B base & +2 [-1, +5] & -1 [-4, +3] & -3 [-8, +1] & +6 [+1, +11] & +6 [+1, +11] \\
Gold & d3-nano & -5 [-9, -1] & -12 [-20, -5] & -20 [-30, -11] & -11 [-21, -1] & -24 [-34, -15] \\
 & Jev & +1 [+0, +2] & +4 [+1, +7] & -41 [-55, -28] & -- & -- \\
 & Luna Decisions & -1 [-5, +3] & +1 [-3, +5] & -13 [-20, -6] & +9 [+4, +15] & -20 [-32, -9] \\
 & Luna gen. none & +1 [-1, +4] & -1 [-4, +1] & -21 [-28, -13] & -2 [-7, +3] & -34 [-44, -24] \\
 & Luna gen. low & -1 [-4, +1] & +0 [-2, +2] & -6 [-11, -2] & -3 [-6, +0] & -3 [-7, +1] \\
 & Qwen3.5-2B base & +3 [+1, +6] & +1 [-4, +7] & +10 [+2, +18] & +18 [+11, +25] & +18 [+8, +27] \\
\bottomrule
\end{tabular}
\end{table}

Agreement across all six representations ranges from 31\% to 78\% on MOCLIP (Table~\ref{tab:consistency}). Much of this variation follows accuracy. Low-reasoning Luna agrees on 78\% of variants, and all these answers are correct. Independent errors at its per-representation accuracy would already give 73\% all-correct agreement. Its disagreements occur mainly in sufficiency, with 43 of 96 variants, and compare, with 54 of 192. Only 2 of 144 specification variants disagree. In less accurate models, many consistent answers are wrong: 12 of d3-nano's 37 points of agreement and 9 of Luna Decisions' 45 points. We therefore use paired accuracy differences in Table~\ref{tab:repeff} as the main measure of representation effects.

\begin{table}[t]
\centering\small
\caption{Agreement across equivalent representations (\%, 95\% intervals) on core variants. ``Wrong'' is the consistent-but-wrong share of all-six agreement; ``Indep.'' is the all-correct share expected if errors were independent across representations.}
\label{tab:consistency}
\setlength{\tabcolsep}{4pt}
\begin{tabular}{lrrrrrrrr}
\toprule
& \multicolumn{4}{c}{MOCLIP} & \multicolumn{4}{c}{Gold} \\
\cmidrule(lr){2-5}\cmidrule(lr){6-9}
Model & All-four & All-six & Wrong & Indep. & All-four & All-six & Wrong & Indep. \\
\midrule
d3-nano & 53 [48, 58] & 37 [32, 42] & 12 & 8 & 62 [53, 71] & 46 [38, 54] & 7 & 6 \\
Jev & 75 [71, 79] & -- & -- & -- & 49 [39, 61] & -- & -- & -- \\
Luna Decisions & 59 [55, 63] & 45 [41, 48] & 9 & 15 & 66 [60, 73] & 47 [39, 55] & 1 & 29 \\
Luna gen. none & 57 [52, 62] & 31 [27, 36] & 4 & 17 & 65 [57, 73] & 39 [32, 47] & 1 & 28 \\
Luna gen. low & 84 [81, 88] & 78 [74, 82] & 0 & 73 & 89 [84, 94] & 87 [82, 92] & 0 & 84 \\
Qwen3.5-2B base & 41 [37, 45] & 31 [27, 34] & 15 & 1 & 66 [61, 71] & 40 [33, 47] & 15 & 3 \\
\bottomrule
\end{tabular}
\end{table}

\begin{figure}[t]
\centering
\includegraphics[width=\linewidth]{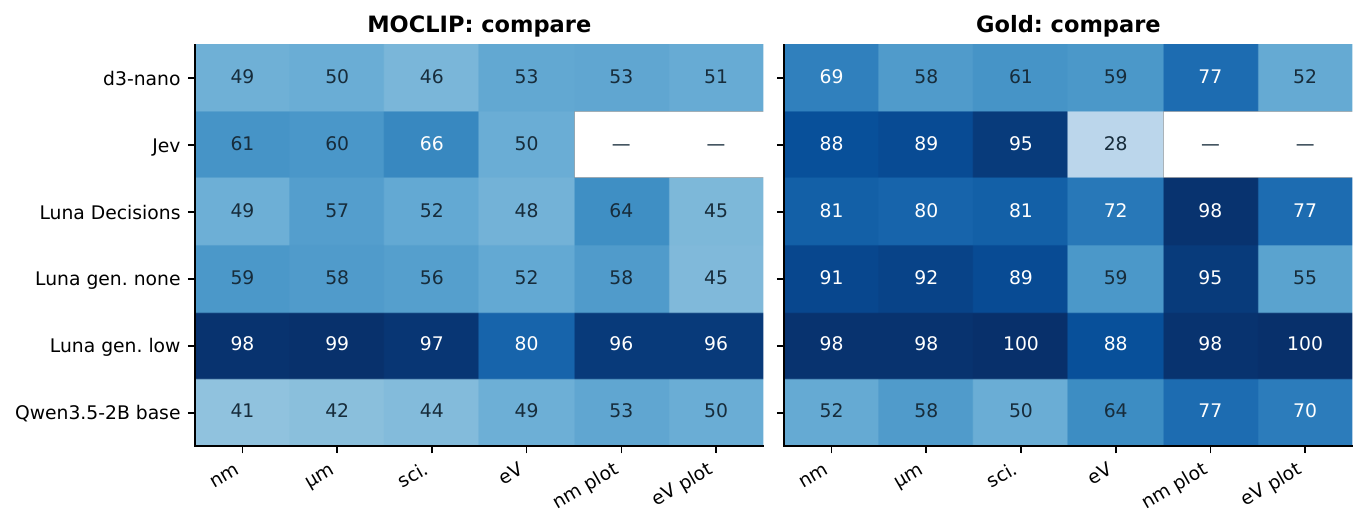}
\caption{Accuracy (\%) on wavelength-worded compare variants in each representation. MOCLIP cells contain 192 requests from 48 instances; gold cells 64 requests from 16 instances. Compare has two options, so 50\% is chance. Jev has no image input; the Qwen3.5-2B base row comes from the supplementary batch.}
\label{fig:compare}
\end{figure}

Comparison questions make direction errors visible (Figure~\ref{fig:compare}). Jev is correct on only 28\% of gold eV-table comparisons. This is well below chance for two options and suggests a systematic reversal of direction. On MOCLIP eV tables, it is at chance, with 50\% accuracy. d3-nano and Luna Decisions are near chance on MOCLIP comparisons in every table format. For these models, the differences between representations mainly reflect a preference for the first spectrum, as discussed below.

Table~\ref{tab:energy} gives all wording comparisons. With the eV evidence unchanged, photon-energy wording improves Jev accuracy by $+16$ on MOCLIP, $+48$ on gold compare, and $+50$ on gold threshold. It also improves low-reasoning Luna on MOCLIP tables by $+20$, Luna Decisions on plots by $+19$ and $+11$, and d3-nano on gold threshold by $+50$ and $+48$. On the last task, d3-nano always chooses ``shorter'' with wavelength wording. Several cells show no detectable change, and others become worse. d3-nano drops from 59\% to 34\% on gold compare tables ($-25$), and low-reasoning Luna drops from 100\% to 90\% on gold threshold tables ($-10$). Matching the question wording to the axis therefore does not help every model or task.

\begin{table}[t]
\centering\small
\caption{Wording effect on identical eV evidence: accuracy with wavelength wording $\to$ photon-energy wording (\%), and the paired difference in points with 95\% interval. Gold has 16 compare and 20 threshold instances.}
\label{tab:energy}
\setlength{\tabcolsep}{2.5pt}
\resizebox{\linewidth}{!}{%
\begin{tabular}{lllllll}
\toprule
& \multicolumn{2}{c}{MOCLIP compare} & \multicolumn{2}{c}{Gold compare} & \multicolumn{2}{c}{Gold threshold} \\
\cmidrule(lr){2-3}\cmidrule(lr){4-5}\cmidrule(lr){6-7}
Model & eV table & eV plot & eV table & eV plot & eV table & eV plot \\
\midrule
d3-nano & 53$\to$52 (-1 [-5, +4]) & 51$\to$54 (+3 [-4, +10]) & 59$\to$34 (-25 [-42, -8]) & 52$\to$62 (+11 [-3, +25]) & 50$\to$100 (+50 [+25, +75]) & 50$\to$98 (+48 [+25, +70]) \\
Jev & 50$\to$66 (+16 [+9, +23]) & -- & 28$\to$77 (+48 [+30, +64]) & -- & 50$\to$100 (+50 [+25, +75]) & -- \\
Luna Decisions & 48$\to$52 (+4 [+1, +7]) & 45$\to$64 (+19 [+9, +28]) & 72$\to$72 (+0 [-9, +8]) & 77$\to$88 (+11 [+2, +20]) & 68$\to$75 (+8 [-12, +28]) & 55$\to$85 (+30 [+0, +60]) \\
Luna gen. none & 52$\to$58 (+6 [-5, +16]) & 45$\to$51 (+6 [-2, +15]) & 59$\to$62 (+3 [-8, +12]) & 55$\to$70 (+16 [+0, +30]) & 72$\to$85 (+12 [-2, +28]) & 60$\to$82 (+22 [+8, +40]) \\
Luna gen. low & 80$\to$99 (+20 [+14, +26]) & 96$\to$96 (+1 [-4, +5]) & 88$\to$95 (+8 [+0, +14]) & 100$\to$98 (-2 [-5, +0]) & 100$\to$90 (-10 [-20, -2]) & 100$\to$98 (-2 [-8, +0]) \\
Qwen3.5-2B base & 49$\to$55 (+5 [-1, +11]) & 50$\to$49 (-1 [-9, +7]) & 64$\to$33 (-31 [-45, -14]) & 70$\to$38 (-33 [-62, -2]) & 50$\to$50 (+0 [+0, +0]) & 50$\to$68 (+18 [+2, +32]) \\
\bottomrule
\end{tabular}}
\end{table}

Models also prefer particular evidence positions. In balanced core comparisons, the first spectrum is correct on half the requests. Yet on MOCLIP tables, Luna Decisions chooses it 82\% of the time, Jev 69\%, Luna gen.\ none 65\%, d3-nano 60\%, and low-reasoning Luna 53\%. Option-order averaging barely changes this preference for d3-nano: it is 60\% without averaging and 58\% with it on MOCLIP, and 69\% and 70\% on gold. Reversing the options does not change which spectrum appears first in the evidence. Jev's result is also consistent with its reported insensitivity to option reordering \citep{pai2026jevbench}. In this case, the preference follows evidence position.

\paragraph{Row-order control.} Compared with nm tables, eV tables reverse the wavelength row order and have uneven coordinate spacing. In the supplementary batch, we sorted each eV table by descending energy to recover the nm row order. We also reran the ascending version on the same day (Table~\ref{tab:evorder}). The reruns reproduced 81--100\% of earlier choices, including 99--100\% for d3-nano and Luna Decisions.

Changing row order helps some tasks and hurts others. Low-reasoning Luna's wavelength-worded comparison accuracy rises from 80\% to 89\%, and Jev's gold comparison accuracy rises from 50\% to 83\%. However, Jev's MOCLIP accuracy with photon-energy wording falls from 66\% to 54\%. Scope discrimination on eV tables also falls for most API configurations: from $+37$ to $+15$ for low-reasoning Luna on MOCLIP and from $+43$ to $+5$ for Jev on gold. MOCLIP middle-gap abstention stays at 0--6\% in both eV row orders, even though low-reasoning Luna abstains on every nm table with that row order. Row order therefore explains part of the eV effect, but not the failure to recognize middle gaps.

\begin{table}[t]
\centering\small
\caption{Row-order control on eV tables (supplementary batch): ascending energy $\to$ descending energy (same row order as nm tables), with paired differences and 95\% intervals. Compare and middle-gap columns are accuracy and abstention rate (\%); scope columns are $\scope$ (points) on eV tables. Gold compare pools both wordings. Middle-gap cells contain 16 instances.}
\label{tab:evorder}
\setlength{\tabcolsep}{2pt}
\resizebox{\linewidth}{!}{%
\begin{tabular}{lllllll}
\toprule
& \multicolumn{4}{c}{MOCLIP} & \multicolumn{2}{c}{Gold} \\
\cmidrule(lr){2-5}\cmidrule(lr){6-7}
Model & Compare, wavelength & Compare, energy & Middle-gap abstention & Scope $\scope$ & Compare & Scope $\scope$ \\
\midrule
d3-nano & 52$\to$54 (+2 [-3, +7]) & 52$\to$49 (-3 [-7, +1]) & 0$\to$0 (+0 [+0, +0]) & +0$\to$+0 (+0 [+0, +0]) & 48$\to$56 (+9 [+2, +16]) & +0$\to$+0 (+0 [+0, +0]) \\
Jev & 51$\to$54 (+4 [-3, +10]) & 66$\to$54 (-12 [-20, -6]) & 0$\to$0 (+0 [+0, +0]) & +7$\to$+7 (+0 [-9, +9]) & 50$\to$83 (+33 [+20, +44]) & +43$\to$+5 (-38 [-53, -23]) \\
Luna Decisions & 48$\to$51 (+3 [-2, +8]) & 52$\to$50 (-2 [-7, +4]) & 0$\to$0 (+0 [+0, +0]) & +26$\to$+9 (-17 [-26, -7]) & 72$\to$68 (-4 [-11, +2]) & +40$\to$+27 (-13 [-30, +3]) \\
Luna gen. none & 49$\to$51 (+2 [-7, +10]) & 56$\to$55 (-1 [-10, +8]) & 0$\to$0 (+0 [+0, +0]) & +19$\to$+2 (-17 [-28, -6]) & 59$\to$74 (+16 [+7, +24]) & +58$\to$+67 (+8 [+2, +15]) \\
Luna gen. low & 80$\to$89 (+9 [+3, +17]) & 100$\to$100 (+0 [+0, +0]) & 0$\to$6 (+6 [+0, +19]) & +37$\to$+15 (-22 [-35, -9]) & 93$\to$97 (+4 [-0, +8]) & +75$\to$+72 (-3 [-10, +3]) \\
Qwen3.5-2B base & 49$\to$55 (+5 [+0, +10]) & 55$\to$51 (-4 [-9, +1]) & 0$\to$0 (+0 [+0, +0]) & +0$\to$+0 (+0 [+0, +0]) & 48$\to$52 (+3 [-1, +7]) & +0$\to$+0 (+0 [+0, +0]) \\
\bottomrule
\end{tabular}}
\end{table}

\paragraph{Same-backbone reference.} With option-letter readout, the Qwen3.5-2B base model is less accurate than d3-nano on MOCLIP: 40\% versus 56\% on tables and 46\% versus 60\% on plots. It is also less accurate on gold tables, at 44\% versus 64\% (Table~\ref{tab:accuracy}). It chooses the first spectrum more often: 72\% of balanced table comparisons on MOCLIP and 82\% on gold, compared with 60\% and 69\% for d3-nano. Like d3-nano, it never abstains in the scope sets, giving $\scope=0$ in every representation (Table~\ref{tab:scope}). Option letters account for 97--99\% of its next-token probability, so the readout is well formed. Both models become worse on gold eV-table comparisons with photon-energy wording: base-model accuracy falls from 64\% to 33\%, and d3-nano from 59\% to 34\%. Decision fine-tuning improves accuracy and reduces the evidence-order preference here, but does not improve abstention when the required evidence is missing.

\subsection{Abstention when relevant evidence is missing}
In the main MOCLIP sufficiency set, we remove a middle segment containing the true dip. d3-nano, Jev, and Luna Decisions never answer ``undetermined'' in any representation. Low-reasoning Luna abstains on 100\% of nm tables, 94\% of $\mu$m tables, and 62\% of scientific-notation tables. It abstains on 0\% of eV tables and only 0--19\% of plots. Each cell has 16 instances.

The scope sets compare these failures with deletions that leave the question answerable (Table~\ref{tab:scope}). Unadapted API configurations almost never abstain on outside cuts, and their accuracy changes by at most seven points. They also discriminate poorly between inside and outside cuts on MOCLIP. Table discrimination is $+14$ for Jev, $+18$ for Luna Decisions, $+19$ for Luna gen.\ none, and $+55$ for low-reasoning Luna. Plot discrimination is at most $+16$. Base d3-nano never abstains. On gold, Jev, Luna Decisions, and Luna gen.\ none choose ``undetermined'' on only 0--5\% of middle-gap requests. Low-reasoning Luna reaches 20--75\% on tables and 0\% on plots.

After an inside cut, a common wrong answer is the band containing the visible extremum. Across deletion types in the main MOCLIP sufficiency set, 54--81\% of d3-nano responses and 31--54\% of Jev responses select this band, depending on representation. In specification tasks, 75--100\% of errors select the candidate that is deep enough but outside the required window.

\begin{table}[t]
\centering\small
\caption{Scope sets: full-view accuracy (Full, \%), scope discrimination $\scope$ and outside-cut accuracy change $\outside$ (points). Ideal: high Full, $\scope=+100$, $\outside=0$. Tables average four representations, plots two. ftC is the LoRA arm trained with answerable truncations. Intervals are in Appendix~\ref{app:tables}.}
\label{tab:scope}
\setlength{\tabcolsep}{4pt}
\begin{tabular}{lrrrrrrrrr}
\toprule
& \multicolumn{3}{c}{MOCLIP tables} & \multicolumn{3}{c}{MOCLIP plots} & \multicolumn{3}{c}{Gold tables} \\
\cmidrule(lr){2-4}\cmidrule(lr){5-7}\cmidrule(lr){8-10}
Model & Full & $\scope$ & $\outside$ & Full & $\scope$ & $\outside$ & Full & $\scope$ & $\outside$ \\
\midrule
d3-nano & 69 & +0 & -3 & 67 & +3 & +1 & 79 & +0 & +5 \\
Jev & 99 & +14 & +1 & -- & -- & -- & 100 & +45 & +0 \\
Luna Decisions & 88 & +18 & -1 & 90 & +4 & +7 & 68 & +42 & +4 \\
Luna gen. none & 95 & +19 & -2 & 88 & +4 & -1 & 96 & +65 & +3 \\
Luna gen. low & 98 & +55 & -1 & 99 & +16 & +1 & 98 & +82 & +2 \\
Qwen3.5-2B base & 44 & +0 & +13 & 56 & +0 & -3 & 50 & +0 & +0 \\
ftC seed 0 & 98 & +96 & +0 & 72 & +86 & +12 & 67 & +70 & +9 \\
ftC seed 1 & 94 & +94 & +0 & 72 & +60 & -8 & 52 & +70 & -2 \\
ftC seed 2 & 100 & +90 & -1 & 81 & +69 & -13 & 65 & +73 & +19 \\
\bottomrule
\end{tabular}
\end{table}

\subsection{Adaptation on table tasks and transfer to other settings}
\label{sec:training}
We adapt only d3-nano. LoRA modules with rank 16, scale 32, and dropout 0.05 are added to the language-model projection layers. The vision tower, backbone weights, readout, and temperature remain frozen. We use AdamW with learning rate $10^{-4}$ and no weight decay, accumulate gradients over eight examples, clip gradients at 1.0, and minimize cross-entropy over the candidate readout. Training uses the evaluation templates, with structures excluded from every test set. It includes neither plots nor specification tasks. A hash of the instance identifier assigns about 10\% of instances to development (Table~\ref{tab:arms}). Some structures are reused when adding answerable truncations or original-task examples. This leaves 8 structures shared between training and development in ftC and 11 in ftD. Development accuracy in these arms is therefore not measured on fully separate structures.

\begin{table}[t]
\centering\small
\caption{LoRA arms (three seeds each). Train/Dev count rendered requests. Selected-epoch arms run at most nine epochs, keep the best development accuracy (loss breaks ties) and stop after two epochs without improvement.}
\label{tab:arms}
\begin{tabular}{llrrl}
\toprule
Arm & Training data and representations & Train & Dev & Duration \\
\midrule
ftA & Original tasks, nm tables & 632 & 80 & 2 epochs \\
ftAc & As ftA & 632 & 80 & Selected epoch \\
ftBm & Original tasks, four tables, sampled to ftA size & 632 & 372 & 2 epochs \\
ftBmc & As ftBm & 632 & 372 & Selected epoch \\
ftB & Original tasks, four tables & 2,956 & 372 & 2 epochs \\
ftC & ftB data + 72 answerable/unanswerable scope instances & 3,736 & 456 & 2 epochs \\
ftD & ftB data + 62 further original-task instances & 3,696 & 512 & 2 epochs \\
ftCc, ftDc & As ftC, ftD & 3,736; 3,696 & 456; 512 & Selected epoch \\
\bottomrule
\end{tabular}
\end{table}

\paragraph{Training duration.} After two epochs on nm tables only, ftA remains at chance on compare, with development accuracy of 0.675--0.700. Training the same data to the selected epoch (ftAc) raises development accuracy to 0.963--1.000 and compare accuracy on nm tables to 85--100\%. The two-epoch model was therefore under-trained. Longer training also reduces variation across seeds in the sample-matched four-table arm. Table compare accuracy changes from 63/97/81\% for ftBm to 99/94/98\% for ftBmc.

\begin{figure}[t]
\centering
\includegraphics[width=\linewidth]{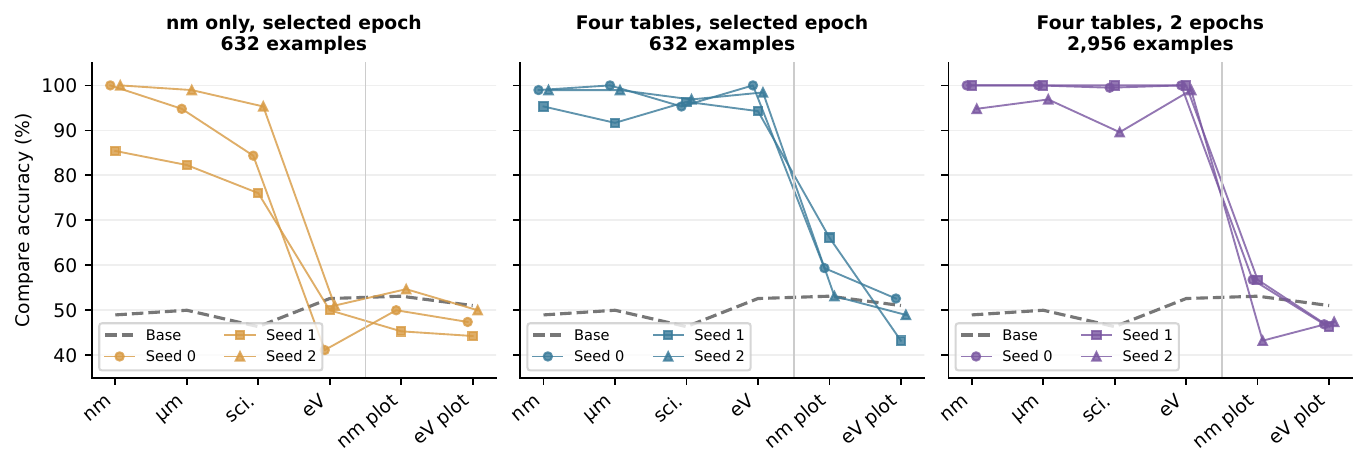}
\caption{MOCLIP compare accuracy (core variants) after adaptation; markers identify seeds and dashed curves show the unadapted model.}
\label{fig:lora}
\end{figure}

\paragraph{Representation coverage.} ftAc reaches 82--99\% on $\mu$m tables and 76--95\% on scientific-notation tables, but only 41--51\% on eV tables (Figure~\ref{fig:lora}). With the same number of examples, the four-table arm ftBmc reaches 94--100\% on eV tables. This is close to ftB, which uses 4.7 times as many examples and reaches 99--100\%. In this comparison, including the representation in training matters more for eV transfer than the total example count. The four-table pool also includes energy-worded variants, so representation and wording coverage are not fully separated.

\paragraph{Transfer to other tasks and data.} Plot comparison remains at 43--66\% for nearly every adapted model, compared with 52\% for the base model. The exception is ftC seed 0, with 72\% on nm plots and 65\% on eV plots. Specification is absent from training. Its accuracy ranges from 35\% to 96\% in the four-table arms and falls to 31--35\% in the nm-only arms, compared with 60\% for the base model.

On gold, table threshold accuracy drops from 82\% to 49--81\% across ftAc, ftBmc, ftB, and ftC; ftA reaches 74--75\%. We check whether the adapters have learned to find a minimum instead of a maximum. Of the 27 adapters, 23 agree with this reversed rule on only 5--26\% of core table requests where the rules differ. The remaining adapters reach 34--66\%: ftB seed 2, ftC seed 1, and ftCc seeds 0 and 1. A common reversal of the rule therefore cannot explain the decline. Converting gold absorbance to transmission $T=10^{-A}$ and asking for a dip also does not recover transfer (Appendix~\ref{app:gold}). The supplementary arms ftD, ftCc, and ftDc show similar results: plot comparison of 51--63\%, specification accuracy of 35--87\%, and gold threshold accuracy of 48--81\%. Table~\ref{tab:allseeds} gives each seed for all 27 adapters.

\subsection{Learning to abstain on every truncated input}
\begin{figure}[t]
\centering
\includegraphics[width=\linewidth]{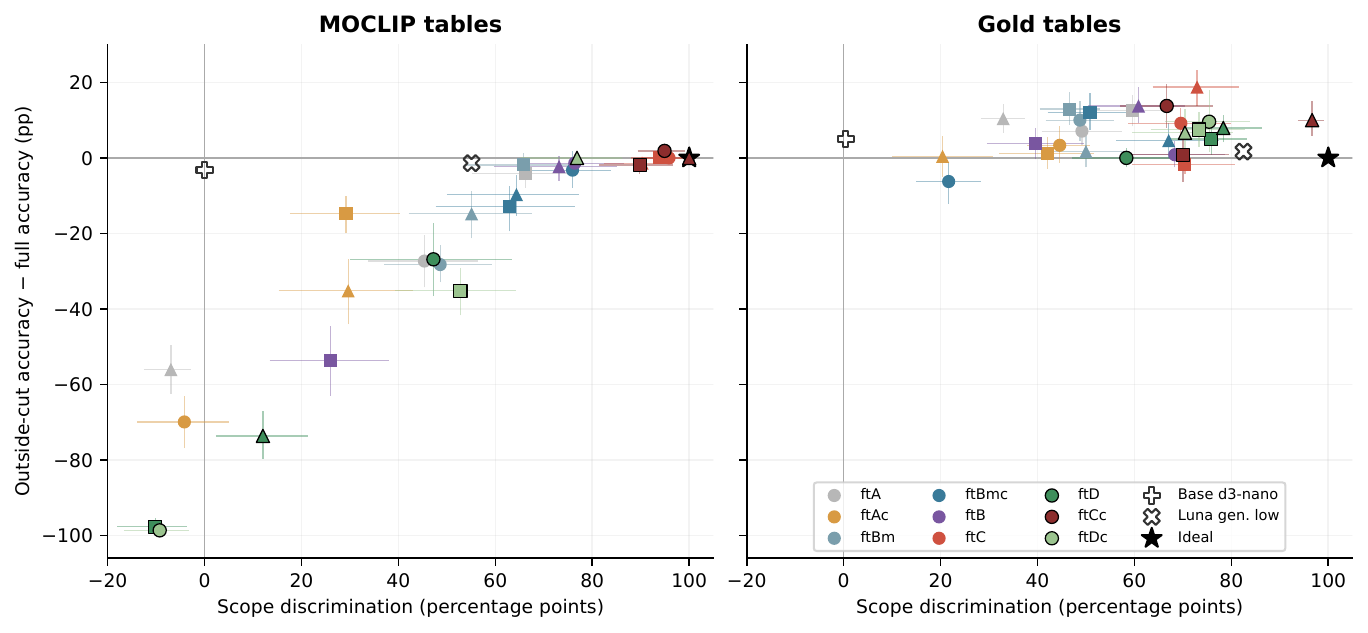}
\caption{Scope behavior of every LoRA seed on tables, with 95\% instance-bootstrap intervals. Circles, squares and triangles mark seeds 0, 1 and 2; outlined markers are the supplementary arms (ftD, ftCc, ftDc). Open markers show unadapted d3-nano and low-reasoning Luna; the star is the ideal $(100,0)$.}
\label{fig:shortcut}
\end{figure}

Adapted models reach 88--100\% table accuracy on the main MOCLIP sufficiency task, compared with 39\% for the base model. However, the scope sets show that some gains come from abstaining on any truncated input (Figure~\ref{fig:shortcut}). Several seeds lose 54--70 points on outside cuts, even though only irrelevant samples are removed: ftA seed 2, ftB seed 1, and ftAc seed 0. Longer training on the original labels can worsen this behavior. For nm-only training, mean discrimination falls from $+35$ in ftA to $+18$ in ftAc, while the mean outside-cut change falls from $-29$ to $-40$.

Adding scope instances in ftC gives MOCLIP table discrimination of $+96$, $+94$, and $+90$ across the three seeds. Outside-cut accuracy changes are $0$, $0$, and $-1$. There is some transfer to plots despite the absence of images in training: discrimination is $+86$, $+60$, and $+69$. Plot outside-cut changes are less stable, at $+12$, $-8$, and $-13$. On gold tables, ftC discrimination is $+70$ to $+73$. This is close to the best earlier seeds, $+68$ for ftB and $+67$ for ftBmc, but below low-reasoning Luna at $+82$. Full-view gold accuracy is only 52--67\%, compared with 79\% for the base model. Also, ftC seed 1 reaches only 0.748 development accuracy and 55\% on MOCLIP compare tables, despite its high scope discrimination.

\paragraph{Added-data control.} ftC adds new examples and increases the number of training updates. To separate these effects, ftD adds 62 instances of the original task types. It has 3,696 training requests, close to ftC's 3,736, and uses the same two-epoch training. All ftD seeds learn the source tasks, reaching 0.99--1.00 development accuracy and 98--99\% on MOCLIP compare tables. However, they still abstain too often on answerable truncations. MOCLIP table discrimination is $+47$, $-10$, and $+12$, with outside-cut changes of $-27$, $-98$, and $-74$. On gold tables, ftD matches or exceeds ftC: discrimination is $+58$ to $+78$, outside-cut changes are $0$ to $+8$, and full-view accuracy is 51--70\%. Adding answerable-truncation examples therefore gives more consistent MOCLIP scope performance in these runs. On gold, a similar amount of original-task data gives comparable scope behavior.

\paragraph{Training to the selected epoch.} With ftC's data and the selected-epoch rule, ftCc stops at epochs 2, 2, and 5. Development accuracy is 0.99--1.00, and MOCLIP compare-table accuracy is 96--100\%, resolving the under-training seen in ftC seed 1. Scope performance also remains good. MOCLIP table discrimination is $+95$, $+90$, and $+100$, with outside-cut changes of $+2$, $-2$, and $0$. Plot discrimination is lower than in ftC, at $+56$ to $+64$, while gold table discrimination ranges from $+67$ to $+97$.

The matched control ftDc uses the same training rule on ftD's data and still varies by seed. MOCLIP table discrimination is $-9$, $+53$, and $+77$, with outside-cut changes of $-99$, $-35$, and $0$. Plot discrimination is $-5$ to $+42$. Across both training durations, all six seeds trained with answerable truncations avoid the shortcut. Five of the six seeds trained with a similar amount of original-task data lose 27--99 points on outside cuts. On gold tables, the converged arms again give similar results: $+67$ to $+97$ for ftCc and $+70$ to $+75$ for ftDc.

\subsection{Comparison of reasoning and interface settings}
On matched core requests, low-reasoning generative Luna exceeds Luna Decisions by 28 points on MOCLIP and 15 on gold (Table~\ref{tab:interface}). Luna gen.\ none differs by only $+3$ and $0$. It is better on tables and worse on plots (Table~\ref{tab:repeff}). Without reasoning, high-confidence errors are less frequent on the decision route: 6\% versus 31\% of MOCLIP requests, and 3\% versus 18\% on gold. With low reasoning, the generative route reduces these fractions to 6\% and 3\%. Low-reasoning generation costs about twice as much as the decision route in these runs: \$0.405 versus \$0.208 for 3,552 MOCLIP requests. It takes about 3\,s per request, compared with 0.3\,s.

\begin{table}[t]
\centering\small
\caption{Matched Luna configurations on core requests: accuracy difference from Luna Decisions (points) and high-confidence error fraction (\% of all requests with confidence $\ge0.9$ and a wrong answer), with 95\% intervals.}
\label{tab:interface}
\begin{tabular}{lrrrrr}
\toprule
& \multicolumn{2}{c}{Difference from Decisions} & \multicolumn{3}{c}{High-confidence error (\%)} \\
\cmidrule(lr){2-3}\cmidrule(lr){4-6}
Dataset & Gen.\ none & Gen.\ low & Decisions & Gen.\ none & Gen.\ low \\
\midrule
MOCLIP & +3 [+1, +4] & +28 [+26, +31] & 6 [5, 7] & 31 [28, 34] & 6 [4, 7] \\
Gold & +0 [-2, +3] & +15 [+12, +19] & 3 [2, 5] & 18 [15, 22] & 3 [2, 4] \\
\bottomrule
\end{tabular}
\end{table}

\section{Discussion}
\paragraph{Coordinate interpretation.} Most models change little when only the wavelength number format changes. Energy coordinates cause larger differences. Asking directly about photon energy helps some models and hurts others on the same evidence. Jev's below-chance result on gold eV tables suggests a systematic direction error. To assess this behavior, accuracy and paired representation differences need to be considered together with controls where the correct answer changes. Agreement alone adds little because it largely follows accuracy.

\paragraph{Evidence needed for the question.} Removing samples outside the query interval leaves the answer available. Removing the relevant extremum does not. Unadapted models often answer both, while some adapted models abstain on both. A high abstention rate on unanswerable deletions alone cannot distinguish these behaviors. Similar problems have been reported in retrieval-augmented question answering \citep{joren2025sufficient,abdumalikov2024answerability}. In our experiments, answerable truncations remove the shortcut in every training seed. A similar amount of extra original-task data, with or without longer training, does so in only one of six seeds.

\paragraph{Limited transfer after adaptation.} LoRA improves comparison on table representations included in training. Adding answerable truncations also improves scope behavior on the source dataset, with some transfer to plots. Plot comparison remains weak, performance on an unseen task varies widely, and accuracy on a second optical dataset falls. These gains are limited to the tested templates, representations, and data. They do not establish that the model has learned a general physical rule.

\section{Limitations}
Both datasets cover a narrow range of optical systems, and we select spectra with clear extrema. MOCLIP's 10\,nm sampling does not support claims about narrow resonances. The gold dataset contains repeated nominal recipes across batches, so instance-bootstrap intervals may understate uncertainty for a new experimental campaign.

The tasks use fixed multiple-choice templates. Each scope set has two query intervals, and their boundaries are partly associated with answer bands. We vary plot style, noise, and phrasing only through the stated controls. Only sufficiency tasks offer an abstention option.

The adaptation study uses one base model, a frozen vision tower, and three seeds per arm. The arms differ in training updates, and two epochs leave ftC seed 1 under-trained. The supplementary and original arms also use different GPUs, so hardware differs in the ftD--ftC comparison. We cannot verify whether service routes share the same internal model, and we did not evaluate the native OpenAI decision endpoint. API models may change over time, and local runs use different GPUs across rounds. The study developed through sequential pilots and controls; all comparisons are exploratory.

\section{Conclusion}
For the tested models, equivalent optical data often lead to different decisions, and removing required evidence seldom leads to abstention. Paired representation tests show errors that average accuracy can hide. Testing both answerable and unanswerable partial spectra also reveals whether a model abstains simply because data are missing. Task-specific adaptation improves the cases covered by training, but transfer to other representations, tasks, and optical data remains limited.

\paragraph{Data and code availability.}
The measurements are available from Figshare \citep{rodionov2026moclip} and Zenodo \citep{lachowski2022gold} under CC BY 4.0. The task definitions, per-response results, training-log summaries and a script that regenerates all result tables and figures are available at \url{https://github.com/zwang-devv/spectra-decision-audit}. The rendering and model-calling code, raw model outputs and adapter weights are available from the authors on request.

\paragraph{Ethics statement.}
The study uses public, licensed physical measurements and involves no human subjects or personal data.

\paragraph{Declarations.}
Funding: this work received no funding. Competing interests: the authors declare none. Author contributions: Z.W. designed and conducted the study and wrote the manuscript; G.Z. and X.C. contributed ideas and reviewed the manuscript.

\FloatBarrier
\bibliographystyle{plainnat}
\bibliography{references}

\clearpage
\appendix
\section{Task and execution details}
MOCLIP band labels are 450--605, 605--645, 645--685 and 685--730\,nm, with boundaries between sampled wavelengths. Specification windows are 605--665, 625--685 and 645--705\,nm; positive candidates lie at least 15\,nm inside their window and location distractors at least 30\,nm outside; positive and location-distractor dips are at least 0.05 below the depth threshold and shallow distractors at least 0.10 above it. Main MOCLIP contains 96 extreme, 384 compare (192 energy-worded), 144 specification and 96 sufficiency variants; main gold contains 80 threshold (40 energy-worded), 128 compare (64 energy-worded) and 36 sufficiency variants. Jev therefore receives 2,304 MOCLIP and 664 gold main requests, multimodal models 3,552 and 1,048.

Dataset generation seeds are 2 (MOCLIP main), 3 (gold main), 5 and 6 (scope sets), and 10, 11 and 12 for the three training pools. LoRA targets are \texttt{q\_proj}, \texttt{k\_proj}, \texttt{v\_proj}, \texttt{o\_proj}, \texttt{in\_proj\_qkv}, \texttt{in\_proj\_z}, \texttt{out\_proj}, \texttt{gate\_proj}, \texttt{up\_proj} and \texttt{down\_proj} (15,630,336 trainable parameters). Development assignment uses the SHA-256 of the instance identifier modulo 1,000 (values below 100). Sample-matched arms draw 632 requests without replacement using the optimization seed. No model is selected by a test score. Main d3-nano inference ran on an RTX 2000 Ada and the original training rounds on L4 GPUs; the supplementary batch (Appendix~\ref{app:102e}) ran on an A40. For each model, task and representation the last successful response is used; requests that failed and were retried count once.

\section{Request examples}
\label{app:prompts}
\small
\input{tables/prompt_examples.tex}
\normalsize

\section{Additional tables}
\label{app:tables}
This section gives the complete scope results for unadapted models and the results for every LoRA seed. Confidence intervals use the instance bootstrap described in the main text.

\begin{table}[!ht]
\centering\small
\caption{Scope metrics with 95\% intervals for unadapted models.}
\label{tab:scopebase}
\begin{tabular}{lllrrr}
\toprule
Model & Dataset & Modality & Full (\%) & $\scope$ & $\outside$ \\
\midrule
d3-nano & MOCLIP & table & 69 [61, 76] & +0 [+0, +0] & -3 [-9, +2] \\
d3-nano & MOCLIP & plot & 67 [57, 76] & +3 [+0, +6] & +1 [-8, +10] \\
d3-nano & Gold & table & 79 [73, 85] & +0 [+0, +1] & +5 [+2, +8] \\
d3-nano & Gold & plot & 50 [38, 62] & +9 [+4, +14] & +3 [+0, +8] \\
Jev & MOCLIP & table & 99 [97, 100] & +14 [+8, +21] & +1 [+0, +2] \\
Jev & Gold & table & 100 [100, 100] & +45 [+35, +55] & +0 [+0, +0] \\
Luna Decisions & MOCLIP & table & 88 [83, 94] & +18 [+11, +25] & -1 [-4, +1] \\
Luna Decisions & MOCLIP & plot & 90 [83, 95] & +4 [+1, +7] & +7 [+3, +13] \\
Luna Decisions & Gold & table & 68 [60, 76] & +42 [+33, +52] & +4 [-0, +9] \\
Luna Decisions & Gold & plot & 84 [78, 90] & +37 [+27, +48] & +4 [-3, +12] \\
Luna gen. none & MOCLIP & table & 95 [90, 99] & +19 [+12, +27] & -2 [-5, +1] \\
Luna gen. none & MOCLIP & plot & 88 [80, 95] & +4 [+1, +7] & -1 [-6, +4] \\
Luna gen. none & Gold & table & 96 [93, 98] & +65 [+54, +77] & +3 [+1, +6] \\
Luna gen. none & Gold & plot & 99 [98, 100] & +19 [+12, +27] & +1 [+0, +2] \\
Luna gen. low & MOCLIP & table & 98 [96, 100] & +55 [+46, +64] & -1 [-5, +2] \\
Luna gen. low & MOCLIP & plot & 99 [97, 100] & +16 [+8, +24] & +1 [+0, +3] \\
Luna gen. low & Gold & table & 98 [95, 100] & +82 [+75, +89] & +2 [-0, +4] \\
Luna gen. low & Gold & plot & 100 [100, 100] & +62 [+51, +74] & +0 [+0, +0] \\
Qwen3.5-2B base & MOCLIP & table & 44 [36, 51] & +0 [+0, +0] & +13 [+6, +20] \\
Qwen3.5-2B base & MOCLIP & plot & 56 [46, 67] & +0 [+0, +0] & -3 [-10, +6] \\
Qwen3.5-2B base & Gold & table & 50 [38, 62] & +0 [+0, +0] & +0 [+0, +0] \\
Qwen3.5-2B base & Gold & plot & 61 [52, 69] & +0 [+0, +0] & +2 [-12, +15] \\
\bottomrule
\end{tabular}
\end{table}

\FloatBarrier
\begin{table}[!ht]
\centering\small
\caption{Every LoRA seed. Accuracies (\%) on core variants: MOCLIP compare tables (CT) and plots (CP), specification tables (ST), gold threshold tables (GT). Scope columns on tables: full-view accuracy (F, \%), $\scope$ and $\outside$ (points).}
\label{tab:allseeds}
\setlength{\tabcolsep}{4pt}
\begin{tabular}{llrrrrrrrrrr}
\toprule
& & \multicolumn{3}{c}{MOCLIP main} & Gold & \multicolumn{3}{c}{MOCLIP scope} & \multicolumn{3}{c}{Gold scope} \\
\cmidrule(lr){3-5}\cmidrule(lr){7-9}\cmidrule(lr){10-12}
Arm & Seed & CT & CP & ST & GT & F & $\scope$ & $\outside$ & F & $\scope$ & $\outside$ \\
\midrule
ftA & 0 & 51 & 51 & 31 & 75 & 85 & +45 & -27 & 67 & +49 & +7 \\
ftA & 1 & 50 & 50 & 33 & 75 & 87 & +66 & -4 & 67 & +60 & +12 \\
ftA & 2 & 51 & 50 & 34 & 74 & 56 & -7 & -56 & 26 & +33 & +10 \\
ftAc & 0 & 80 & 49 & 32 & 63 & 78 & -4 & -70 & 44 & +45 & +3 \\
ftAc & 1 & 73 & 45 & 35 & 73 & 87 & +29 & -15 & 73 & +42 & +1 \\
ftAc & 2 & 86 & 52 & 31 & 54 & 91 & +30 & -35 & 44 & +20 & +0 \\
ftBm & 0 & 63 & 51 & 37 & 68 & 96 & +49 & -28 & 46 & +49 & +10 \\
ftBm & 1 & 97 & 62 & 73 & 73 & 93 & +66 & -2 & 44 & +47 & +13 \\
ftBm & 2 & 81 & 48 & 60 & 77 & 84 & +55 & -15 & 62 & +50 & +2 \\
ftBmc & 0 & 99 & 56 & 73 & 59 & 90 & +76 & -3 & 30 & +22 & -6 \\
ftBmc & 1 & 94 & 55 & 84 & 79 & 92 & +63 & -13 & 35 & +51 & +12 \\
ftBmc & 2 & 98 & 51 & 45 & 49 & 90 & +64 & -10 & 56 & +67 & +5 \\
ftB & 0 & 100 & 52 & 40 & 50 & 88 & +76 & -1 & 58 & +68 & +1 \\
ftB & 1 & 100 & 52 & 60 & 65 & 95 & +26 & -54 & 40 & +40 & +4 \\
ftB & 2 & 95 & 45 & 55 & 62 & 91 & +73 & -2 & 52 & +61 & +14 \\
ftC & 0 & 99 & 69 & 70 & 64 & 98 & +96 & +0 & 67 & +70 & +9 \\
ftC & 1 & 55 & 49 & 35 & 55 & 94 & +94 & +0 & 52 & +70 & -2 \\
ftC & 2 & 97 & 57 & 96 & 81 & 100 & +90 & -1 & 65 & +73 & +19 \\
ftD & 0 & 98 & 52 & 35 & 49 & 98 & +47 & -27 & 51 & +58 & +0 \\
ftD & 1 & 99 & 59 & 81 & 57 & 98 & -10 & -98 & 66 & +76 & +5 \\
ftD & 2 & 99 & 57 & 80 & 66 & 94 & +12 & -74 & 70 & +78 & +8 \\
ftCc & 0 & 99 & 52 & 68 & 81 & 97 & +95 & +2 & 64 & +67 & +14 \\
ftCc & 1 & 96 & 51 & 48 & 62 & 95 & +90 & -2 & 52 & +70 & +1 \\
ftCc & 2 & 100 & 58 & 87 & 80 & 98 & +100 & +0 & 70 & +97 & +10 \\
ftDc & 0 & 100 & 54 & 40 & 49 & 99 & -9 & -99 & 49 & +75 & +10 \\
ftDc & 1 & 100 & 63 & 84 & 69 & 98 & +53 & -35 & 66 & +73 & +8 \\
ftDc & 2 & 100 & 60 & 58 & 48 & 100 & +77 & +0 & 57 & +70 & +7 \\
\bottomrule
\end{tabular}
\end{table}

\FloatBarrier
\small
\begin{longtable}{ll ll rrr}
\caption{Per-seed scope metrics with 95\% instance-bootstrap intervals.}\label{tab:scopeintervals}\\
\toprule
Arm & Seed & Dataset & Modality & Full (\%) & $\scope$ & $\outside$ \\
\midrule\endfirsthead
\multicolumn{7}{l}{\tablename\ \thetable\ continued}\\
\toprule Arm & Seed & Dataset & Modality & Full (\%) & $\scope$ & $\outside$ \\
\midrule\endhead
\bottomrule\endfoot
ftA & 0 & MOCLIP & table & 85 [78, 91] & +45 [+34, +56] & -27 [-34, -20] \\
ftA & 0 & MOCLIP & plot & 76 [69, 82] & +1 [-10, +12] & -52 [-60, -43] \\
ftA & 0 & Gold & table & 67 [62, 73] & +49 [+41, +57] & +7 [+3, +11] \\
ftA & 0 & Gold & plot & 52 [41, 65] & +53 [+42, +65] & +0 [-4, +4] \\
ftA & 1 & MOCLIP & table & 87 [78, 94] & +66 [+55, +76] & -4 [-8, -0] \\
ftA & 1 & MOCLIP & plot & 77 [68, 85] & +35 [+21, +48] & -19 [-28, -12] \\
ftA & 1 & Gold & table & 67 [63, 70] & +60 [+51, +68] & +12 [+8, +17] \\
ftA & 1 & Gold & plot & 50 [38, 62] & +43 [+33, +52] & -10 [-15, -5] \\
ftA & 2 & MOCLIP & table & 56 [50, 63] & -7 [-12, -3] & -56 [-62, -50] \\
ftA & 2 & MOCLIP & plot & 29 [21, 36] & -4 [-7, -1] & -29 [-36, -21] \\
ftA & 2 & Gold & table & 26 [24, 29] & +33 [+28, +38] & +10 [+7, +14] \\
ftA & 2 & Gold & plot & 18 [12, 23] & +9 [+4, +15] & -8 [-16, -1] \\
ftAc & 0 & MOCLIP & table & 78 [74, 83] & -4 [-14, +5] & -70 [-77, -63] \\
ftAc & 0 & MOCLIP & plot & 64 [54, 73] & -12 [-20, -5] & -61 [-71, -51] \\
ftAc & 0 & Gold & table & 44 [40, 48] & +45 [+38, +51] & +3 [-1, +8] \\
ftAc & 0 & Gold & plot & 44 [32, 55] & +38 [+28, +47] & +1 [-8, +8] \\
ftAc & 1 & MOCLIP & table & 87 [78, 94] & +29 [+18, +40] & -15 [-20, -10] \\
ftAc & 1 & MOCLIP & plot & 69 [63, 76] & +18 [+3, +32] & -12 [-21, -2] \\
ftAc & 1 & Gold & table & 73 [66, 80] & +42 [+32, +52] & +1 [-3, +5] \\
ftAc & 1 & Gold & plot & 63 [53, 73] & +30 [+18, +42] & -8 [-12, -2] \\
ftAc & 2 & MOCLIP & table & 91 [88, 94] & +30 [+15, +43] & -35 [-44, -27] \\
ftAc & 2 & MOCLIP & plot & 77 [69, 84] & +1 [-12, +14] & -45 [-54, -37] \\
ftAc & 2 & Gold & table & 44 [38, 50] & +20 [+10, +31] & +0 [-5, +6] \\
ftAc & 2 & Gold & plot & 56 [44, 68] & +36 [+26, +47] & -1 [-4, +2] \\
ftBm & 0 & MOCLIP & table & 96 [91, 99] & +49 [+37, +59] & -28 [-33, -23] \\
ftBm & 0 & MOCLIP & plot & 80 [71, 87] & +21 [+10, +31] & -47 [-54, -41] \\
ftBm & 0 & Gold & table & 46 [39, 54] & +49 [+42, +56] & +10 [+5, +15] \\
ftBm & 0 & Gold & plot & 28 [19, 37] & +27 [+19, +35] & +4 [-2, +11] \\
ftBm & 1 & MOCLIP & table & 93 [89, 96] & +66 [+54, +77] & -2 [-5, +1] \\
ftBm & 1 & MOCLIP & plot & 69 [61, 78] & +50 [+36, +63] & -5 [-13, +4] \\
ftBm & 1 & Gold & table & 44 [40, 48] & +47 [+40, +53] & +13 [+9, +18] \\
ftBm & 1 & Gold & plot & 53 [44, 62] & +36 [+23, +48] & +9 [+2, +16] \\
ftBm & 2 & MOCLIP & table & 84 [75, 92] & +55 [+42, +68] & -15 [-21, -9] \\
ftBm & 2 & MOCLIP & plot & 78 [68, 86] & +28 [+14, +41] & -26 [-34, -18] \\
ftBm & 2 & Gold & table & 62 [57, 68] & +50 [+41, +59] & +2 [-2, +6] \\
ftBm & 2 & Gold & plot & 49 [37, 62] & +47 [+37, +57] & +2 [+0, +6] \\
ftBmc & 0 & MOCLIP & table & 90 [84, 96] & +76 [+68, +84] & -3 [-8, +2] \\
ftBmc & 0 & MOCLIP & plot & 72 [63, 81] & +37 [+24, +48] & -14 [-22, -6] \\
ftBmc & 0 & Gold & table & 30 [24, 37] & +22 [+15, +28] & -6 [-12, -1] \\
ftBmc & 0 & Gold & plot & 25 [18, 32] & +19 [+10, +28] & +5 [+0, +10] \\
ftBmc & 1 & MOCLIP & table & 92 [85, 97] & +63 [+48, +76] & -13 [-19, -7] \\
ftBmc & 1 & MOCLIP & plot & 77 [67, 86] & +5 [-6, +16] & -46 [-56, -36] \\
ftBmc & 1 & Gold & table & 35 [31, 38] & +51 [+42, +60] & +12 [+8, +17] \\
ftBmc & 1 & Gold & plot & 30 [22, 38] & +42 [+32, +51] & +22 [+16, +29] \\
ftBmc & 2 & MOCLIP & table & 90 [81, 97] & +64 [+50, +77] & -10 [-15, -5] \\
ftBmc & 2 & MOCLIP & plot & 80 [70, 89] & +44 [+31, +58] & -15 [-22, -6] \\
ftBmc & 2 & Gold & table & 56 [46, 67] & +67 [+56, +78] & +5 [+0, +9] \\
ftBmc & 2 & Gold & plot & 52 [40, 64] & +44 [+32, +56] & +3 [+0, +8] \\
ftB & 0 & MOCLIP & table & 88 [79, 94] & +76 [+65, +87] & -1 [-4, +1] \\
ftB & 0 & MOCLIP & plot & 74 [63, 84] & +53 [+43, +63] & -14 [-22, -5] \\
ftB & 0 & Gold & table & 58 [48, 69] & +68 [+57, +79] & +1 [-2, +5] \\
ftB & 0 & Gold & plot & 23 [16, 32] & +20 [+10, +29] & +11 [+5, +17] \\
ftB & 1 & MOCLIP & table & 95 [90, 99] & +26 [+13, +38] & -54 [-63, -44] \\
ftB & 1 & MOCLIP & plot & 81 [71, 89] & +4 [-6, +14] & -54 [-63, -45] \\
ftB & 1 & Gold & table & 40 [36, 45] & +40 [+30, +50] & +4 [-0, +8] \\
ftB & 1 & Gold & plot & 31 [22, 40] & +13 [+4, +23] & +2 [-4, +8] \\
ftB & 2 & MOCLIP & table & 91 [84, 96] & +73 [+60, +85] & -2 [-6, +0] \\
ftB & 2 & MOCLIP & plot & 68 [57, 78] & +52 [+39, +64] & -8 [-16, -1] \\
ftB & 2 & Gold & table & 52 [46, 58] & +61 [+50, +70] & +14 [+9, +19] \\
ftB & 2 & Gold & plot & 52 [40, 64] & +23 [+12, +34] & -4 [-8, -1] \\
ftC & 0 & MOCLIP & table & 98 [96, 100] & +96 [+90, +100] & +0 [+0, +0] \\
ftC & 0 & MOCLIP & plot & 72 [63, 81] & +86 [+79, +93] & +12 [+6, +19] \\
ftC & 0 & Gold & table & 67 [62, 72] & +70 [+59, +80] & +9 [+5, +13] \\
ftC & 0 & Gold & plot & 57 [46, 68] & +29 [+18, +39] & -7 [-14, +0] \\
ftC & 1 & MOCLIP & table & 94 [88, 98] & +94 [+88, +99] & +0 [-1, +1] \\
ftC & 1 & MOCLIP & plot & 72 [63, 81] & +60 [+48, +71] & -8 [-16, -2] \\
ftC & 1 & Gold & table & 52 [46, 57] & +70 [+60, +81] & -2 [-4, +0] \\
ftC & 1 & Gold & plot & 50 [50, 50] & +49 [+38, +60] & +8 [+3, +12] \\
ftC & 2 & MOCLIP & table & 100 [99, 100] & +90 [+82, +97] & -1 [-3, +0] \\
ftC & 2 & MOCLIP & plot & 81 [73, 89] & +69 [+56, +81] & -13 [-20, -6] \\
ftC & 2 & Gold & table & 65 [60, 70] & +73 [+64, +82] & +19 [+14, +23] \\
ftC & 2 & Gold & plot & 52 [41, 64] & +45 [+35, +55] & +5 [-2, +12] \\
ftD & 0 & MOCLIP & table & 98 [95, 100] & +47 [+30, +63] & -27 [-37, -17] \\
ftD & 0 & MOCLIP & plot & 91 [84, 96] & +50 [+36, +64] & -15 [-22, -8] \\
ftD & 0 & Gold & table & 51 [45, 57] & +58 [+47, +69] & +0 [-2, +2] \\
ftD & 0 & Gold & plot & 51 [38, 63] & +46 [+36, +57] & +0 [+0, +0] \\
ftD & 1 & MOCLIP & table & 98 [95, 100] & -10 [-18, -4] & -98 [-100, -95] \\
ftD & 1 & MOCLIP & plot & 69 [59, 78] & -6 [-13, -1] & -69 [-78, -58] \\
ftD & 1 & Gold & table & 66 [58, 74] & +76 [+68, +83] & +5 [+1, +9] \\
ftD & 1 & Gold & plot & 50 [38, 62] & +25 [+15, +36] & -10 [-16, -4] \\
ftD & 2 & MOCLIP & table & 94 [88, 98] & +12 [+2, +21] & -74 [-80, -67] \\
ftD & 2 & MOCLIP & plot & 30 [21, 38] & -3 [-7, +1] & -29 [-37, -20] \\
ftD & 2 & Gold & table & 70 [63, 76] & +78 [+70, +86] & +8 [+4, +11] \\
ftD & 2 & Gold & plot & 40 [29, 51] & +27 [+17, +37] & -1 [-9, +7] \\
ftCc & 0 & MOCLIP & table & 97 [94, 99] & +95 [+89, +99] & +2 [+0, +4] \\
ftCc & 0 & MOCLIP & plot & 67 [58, 75] & +56 [+43, +69] & -6 [-17, +4] \\
ftCc & 0 & Gold & table & 64 [60, 68] & +67 [+57, +76] & +14 [+8, +20] \\
ftCc & 0 & Gold & plot & 64 [54, 74] & +41 [+31, +51] & -1 [-8, +6] \\
ftCc & 1 & MOCLIP & table & 95 [91, 98] & +90 [+81, +97] & -2 [-4, +0] \\
ftCc & 1 & MOCLIP & plot & 83 [73, 93] & +64 [+54, +73] & -2 [-5, +0] \\
ftCc & 1 & Gold & table & 52 [47, 57] & +70 [+59, +80] & +1 [-6, +8] \\
ftCc & 1 & Gold & plot & 58 [53, 63] & +43 [+32, +53] & +7 [+2, +12] \\
ftCc & 2 & MOCLIP & table & 98 [96, 100] & +100 [+100, +100] & +0 [+0, +0] \\
ftCc & 2 & MOCLIP & plot & 50 [42, 59] & +57 [+48, +66] & +7 [+0, +16] \\
ftCc & 2 & Gold & table & 70 [63, 76] & +97 [+94, +99] & +10 [+6, +15] \\
ftCc & 2 & Gold & plot & 39 [29, 49] & +42 [+32, +51] & +11 [+6, +16] \\
ftDc & 0 & MOCLIP & table & 99 [97, 100] & -9 [-17, -3] & -99 [-100, -97] \\
ftDc & 0 & MOCLIP & plot & 40 [34, 45] & -5 [-9, -1] & -40 [-45, -34] \\
ftDc & 0 & Gold & table & 49 [40, 57] & +75 [+66, +84] & +10 [+2, +18] \\
ftDc & 0 & Gold & plot & 32 [23, 41] & +25 [+14, +35] & +4 [-2, +10] \\
ftDc & 1 & MOCLIP & table & 98 [96, 100] & +53 [+39, +64] & -35 [-42, -29] \\
ftDc & 1 & MOCLIP & plot & 85 [76, 94] & +9 [-1, +19] & -58 [-67, -49] \\
ftDc & 1 & Gold & table & 66 [60, 72] & +73 [+63, +83] & +8 [+3, +12] \\
ftDc & 1 & Gold & plot & 52 [41, 64] & +32 [+21, +44] & +8 [+2, +13] \\
ftDc & 2 & MOCLIP & table & 100 [100, 100] & +77 [+65, +86] & +0 [+0, +0] \\
ftDc & 2 & MOCLIP & plot & 84 [76, 91] & +42 [+31, +53] & -3 [-10, +5] \\
ftDc & 2 & Gold & table & 57 [46, 69] & +70 [+60, +80] & +7 [+1, +13] \\
ftDc & 2 & Gold & plot & 50 [38, 62] & +32 [+22, +43] & +1 [-2, +4] \\
\end{longtable}
\normalsize

\section{Gold transmission view}
\label{app:gold}
In an auxiliary set, we use the same gold instances but convert background-subtracted absorbance to $T=10^{-A}$ and ask for the deepest transmission dip. The extremum wavelength is unchanged. The vertical scale, precision, curve shape, and wording all change together. On core table requests, base d3-nano threshold accuracy is 81\%, compared with 82\% for absorbance. ftA seeds reach 68/80/78\%, ftBm reaches 65/50/82\%, and ftB reaches 52/47/71\%. Compare accuracy is 46\% for the base model, 52/50/52\% for ftA, 52/54/40\% for ftBm, and 56/50/44\% for ftB. Using dip wording, as in the source task, does not recover transfer to gold. We did not evaluate later arms on this view.

\section{Supplementary batch}
\label{app:102e}
On 24 October 2026, we added the eV row-order control, the Qwen3.5-2B base reference, and further LoRA arms. The protocol was written before these results were seen. We kept earlier task files, responses, and adapters unchanged and made all comparisons for these analyses within the supplementary batch.

\paragraph{Row order.} The descending-energy table has the same values, wording and options as the original eV table. Every variant that admits an eV table was run in both orders for d3-nano, the Qwen3.5-2B base model, Jev and the three Luna configurations; the oracle is correct on all of them. Same-day ascending reruns reproduced the earlier choices for 99.0--99.6\% of d3-nano requests (on different GPU hardware), 98.8--100\% for Luna Decisions, 94--98\% for Jev, 93--96\% for low-reasoning Luna and 81--95\% for Luna gen.\ none; accuracies changed by at most two points.

\paragraph{Base-model readout.} The base model receives the evidence text, the question, lettered options, and the instruction in Appendix~\ref{app:prompts}, using its chat template with thinking disabled. For image requests, the image precedes the text. We choose the option letter with the highest next-token logit and calculate a softmax over option letters only. On average, these letters carry 97\% of the full next-token probability on MOCLIP and 99\% on gold.

\paragraph{Additional LoRA arms.} The ftD pool adds 6 extreme, 16 compare, and 9 full-range sufficiency instances per polarization, using seed 12. The structure rule is the same as for ftC: additions must be disjoint from every MOCLIP test set but can reuse original training structures. With four table representations, ftD has 3,696 training and 512 development requests, compared with 3,736 and 456 for ftC. ftD, ftCc, and ftDc use the original hyperparameters. The selected epochs are 2, 2, and 5 for ftCc and 9, 2, and 9 for ftDc. Every supplementary adapter reaches development accuracy of 0.99--1.00.

The protocol gives priority in the order ftD $>$ ftCc $>$ ftDc and allows ftDc to be dropped if projected GPU cost exceeds the budget. We initially skipped ftDc based on a nine-epoch estimate, then ran it when early stopping reduced the estimate. All three seeds are reported. A duplicate ftDc seed-2 job was stopped before completing an epoch; the evaluated adapter comes from the stored run. The batch used about 9.3 GPU-hours and \$0.81 in API calls.

\end{document}

%% file: tables/prompt_examples.tex
\noindent\textbf{MOCLIP compare, eV table, wavelength wording} (\texttt{compare-000-orig-longer-wavelength}, \texttt{table\_ev})
\begin{verbatim}
Measured optical transmission of silicon metasurfaces under y-polarized illumination.
  Transmission is the fraction of incident light that passes through (0 = none, 1 = all).

The table lists transmission spectra A and B.

photon_energy_eV,transmission_A,transmission_B
1.698,0.494,0.296
1.722,0.451,0.230
1.746,0.421,0.155
1.771,0.367,0.095
... (25 more rows)

Question: Which spectrum has its deepest transmission dip at the longer wavelength, i.e. which one is
  red-shifted relative to the other?
Options:
- A: Spectrum A
- B: Spectrum B
\end{verbatim}

\noindent\textbf{Same evidence, energy wording} (\texttt{compare-000-orig-longer-axis}, \texttt{table\_ev})
\begin{verbatim}
Measured optical transmission of silicon metasurfaces under y-polarized illumination.
  Transmission is the fraction of incident light that passes through (0 = none, 1 = all).

The table lists transmission spectra A and B.

photon_energy_eV,transmission_A,transmission_B
1.698,0.494,0.296
1.722,0.451,0.230
1.746,0.421,0.155
1.771,0.367,0.095
... (25 more rows)

Question: Which spectrum has its deepest transmission dip at the lower photon energy?
Options:
- A: Spectrum A
- B: Spectrum B
\end{verbatim}

\noindent\textbf{MOCLIP scope task, nm table, inside cut} (\texttt{suffq-009-cut\_in\_hide\_middle}, \texttt{table\_nm})
\begin{verbatim}
Measured optical transmission of silicon metasurfaces under x-polarized illumination.
  Transmission is the fraction of incident light that passes through (0 = none, 1 = all).

The table lists a transmission spectrum.

wavelength_nm,transmission
450,0.410
460,0.463
470,0.495
480,0.526
... (20 more rows)

Question: Considering only the range 450-685 nm, at what wavelength is the transmission lowest? Choose
  the range that contains it. If the data provided are not enough to determine this, choose
  'cannot be determined'.
Options:
- 450-605 nm
- 605-645 nm
- 645-685 nm
- 685-730 nm
- cannot be determined: The data provided are not enough to answer.
\end{verbatim}

\noindent\textbf{Interface wrappers.} Decision interfaces receive the evidence text as \texttt{state}, the question as \texttt{instructions}, and options as a label-to-description map. The generative route appends the options as a bulleted list and the sentence below, with a strict JSON schema requiring \texttt{choice} and \texttt{probabilities}. The Qwen3.5-2B base readout lists lettered options and the final instruction shown; letters are used directly when the option labels are already single letters.
\begin{verbatim}
Generative route: Answer with the option name in `choice`, and give your proba
  bility for every option in `probabilities` (they should sum to 1).
Base readout (band options):
A. 450-605 nm
B. 605-645 nm
C. 645-685 nm
D. 685-730 nm
E. cannot be determined (The data provided are not enough to answer.)
Answer with the letter of the correct option only.
\end{verbatim}

%% file: main.bbl
\begin{thebibliography}{29}
\providecommand{\natexlab}[1]{#1}
\providecommand{\url}[1]{\texttt{#1}}
\expandafter\ifx\csname urlstyle\endcsname\relax
  \providecommand{\doi}[1]{doi: #1}\else
  \providecommand{\doi}{doi: \begingroup \urlstyle{rm}\Url}\fi

\bibitem[Abdumalikov et~al.(2024)Abdumalikov, Minervini, and
  Kementchedjhieva]{abdumalikov2024answerability}
Rustam Abdumalikov, Pasquale Minervini, and Yova Kementchedjhieva.
\newblock Answerability in retrieval-augmented open-domain question answering.
\newblock arXiv:2403.01461, 2024.
\newblock URL \url{https://arxiv.org/abs/2403.01461}.

\bibitem[Alampara et~al.(2025)Alampara, Schilling-Wilhelmi,
  R{\'i}os-Garc{\'i}a, Mandal, Khetarpal, Grover, Krishnan, and
  Jablonka]{alampara2025macbench}
Nawaf Alampara, Mara Schilling-Wilhelmi, Marti{\~n}o R{\'i}os-Garc{\'i}a,
  Indrajeet Mandal, Pranav Khetarpal, Hargun~Singh Grover, N.~M.~Anoop
  Krishnan, and Kevin~Maik Jablonka.
\newblock Probing the limitations of multimodal language models for chemistry
  and materials research.
\newblock \emph{Nature Computational Science}, 5\penalty0 (10):\penalty0
  952--961, 2025.
\newblock \doi{10.1038/s43588-025-00836-3}.

\bibitem[Almeida(2026)]{almeida2026jev}
Diogo Almeida.
\newblock Introducing system one models \& {Jev}.
\newblock TypeSafe AI blog, 2026.
\newblock URL
  \url{https://typesafe.ai/blog/introducing-system-one-models-and-jev}.
\newblock Accessed 24 October 2026.

\bibitem[Azizi et~al.(2026)Azizi, Potraghloo, and Pedram]{azizi2026labels}
Seyedarmin Azizi, Erfan~Baghaei Potraghloo, and Massoud Pedram.
\newblock Labels override definitions in {Jev}-style typed decision models.
\newblock arXiv:2610.02586, 2026.
\newblock URL \url{https://arxiv.org/abs/2610.02586}.

\bibitem[Chen et~al.(2025)Chen, Feng, Wen, Wang, and Fu]{chen2025annular}
Yanlin Chen, Feng Feng, Xin Wen, Ziyuan Wang, and Yiqun Fu.
\newblock Machine learning-driven prediction of optical responses and inverse
  design of annular aperture arrays.
\newblock \emph{Journal of Applied Physics}, 137\penalty0 (24):\penalty0
  243105, 2025.
\newblock \doi{10.1063/5.0276411}.

\bibitem[Deng et~al.(2026)Deng, Fan, Zhang, and Xie]{deng2026jev}
Boyuan Deng, Shuyi Fan, Hongyang Zhang, and Xinhong Xie.
\newblock {Jev} for scientific decisions: Evaluating semantic choices and their
  consequences.
\newblock arXiv:2609.24965, version 3, 2026.
\newblock URL \url{https://arxiv.org/abs/2609.24965v3}.

\bibitem[Deng et~al.(2024)Deng, Sun, He, Sikka, Chen, Ma, Zhang, and
  Mihalcea]{deng2024tables}
Naihao Deng, Zhenjie Sun, Ruiqi He, Aman Sikka, Yulong Chen, Lin Ma, Yue Zhang,
  and Rada Mihalcea.
\newblock Tables as texts or images: Evaluating the table reasoning ability of
  {LLMs} and {MLLMs}.
\newblock In \emph{Findings of the Association for Computational Linguistics:
  ACL 2024}, pages 407--426, 2024.
\newblock \doi{10.18653/v1/2024.findings-acl.23}.

\bibitem[Fu et~al.(2024)Fu, Guo, Khalighinejad, Liu, Dhingra, Yogatama, Jia,
  and Neiswanger]{fu2024isobench}
Deqing Fu, Ruohao Guo, Ghazal Khalighinejad, Ollie Liu, Bhuwan Dhingra, Dani
  Yogatama, Robin Jia, and Willie Neiswanger.
\newblock {IsoBench}: Benchmarking multimodal foundation models on isomorphic
  representations.
\newblock In \emph{First Conference on Language Modeling (COLM)}, 2024.
\newblock URL \url{https://arxiv.org/abs/2404.01266}.
\newblock arXiv:2404.01266.

\bibitem[Geirhos et~al.(2020)Geirhos, Jacobsen, Michaelis, Zemel, Brendel,
  Bethge, and Wichmann]{geirhos2020shortcut}
Robert Geirhos, J{\"o}rn-Henrik Jacobsen, Claudio Michaelis, Richard Zemel,
  Wieland Brendel, Matthias Bethge, and Felix~A. Wichmann.
\newblock Shortcut learning in deep neural networks.
\newblock \emph{Nature Machine Intelligence}, 2:\penalty0 665--673, 2020.
\newblock \doi{10.1038/s42256-020-00257-z}.

\bibitem[Hu et~al.(2021)Hu, Shen, Wallis, Allen-Zhu, Li, Wang, Wang, and
  Chen]{hu2021lora}
Edward~J. Hu, Yelong Shen, Phillip Wallis, Zeyuan Allen-Zhu, Yuanzhi Li, Shean
  Wang, Lu~Wang, and Weizhu Chen.
\newblock {LoRA}: Low-rank adaptation of large language models.
\newblock arXiv:2106.09685, 2021.
\newblock URL \url{https://arxiv.org/abs/2106.09685}.

\bibitem[Joren et~al.(2025)Joren, Zhang, Ferng, Juan, Taly, and
  Rashtchian]{joren2025sufficient}
Hailey Joren, Jianyi Zhang, Chun-Sung Ferng, Da-Cheng Juan, Ankur Taly, and
  Cyrus Rashtchian.
\newblock Sufficient context: A new lens on retrieval augmented generation
  systems.
\newblock In \emph{International Conference on Learning Representations
  (ICLR)}, 2025.
\newblock URL \url{https://arxiv.org/abs/2411.06037}.
\newblock arXiv:2411.06037.

\bibitem[Lachowski et~al.(2022{\natexlab{a}})Lachowski, Vaddi, Naser, Baneyx,
  and Pozzo]{lachowski2022gold}
Kacper~J. Lachowski, Kiran Vaddi, Nada~Y. Naser, Fran{\c c}ois Baneyx, and
  Lilo~D. Pozzo.
\newblock Gold nanoparticles synthesized in the presence of peptides: {UV-Vis}
  spectra, fluorescence, {USAXS}, electron microscopy.
\newblock Zenodo dataset, version v0.1.1, 2022{\natexlab{a}}.
\newblock URL \url{https://zenodo.org/records/6588769}.

\bibitem[Lachowski et~al.(2022{\natexlab{b}})Lachowski, Vaddi, Naser, Baneyx,
  and Pozzo]{lachowski2022multivariate}
Kacper~J. Lachowski, Kiran Vaddi, Nada~Y. Naser, Fran{\c c}ois Baneyx, and
  Lilo~D. Pozzo.
\newblock Multivariate analysis of peptide-driven nucleation and growth of {Au}
  nanoparticles.
\newblock \emph{Digital Discovery}, 1\penalty0 (4):\penalty0 427--439,
  2022{\natexlab{b}}.
\newblock \doi{10.1039/d2dd00017b}.

\bibitem[Lin et~al.(2025)Lin, Liu, Cao, Cao, and Wang]{lin2025transfer}
Yuzhen Lin, Junchen Liu, Yuchen Cao, Yu~Cao, and Ziyuan Wang.
\newblock Transfer learning-enhanced modelling of annular aperture arrays and
  nanohole arrays.
\newblock \emph{Physica Scripta}, 100\penalty0 (3):\penalty0 036003, 2025.
\newblock \doi{10.1088/1402-4896/adae3c}.

\bibitem[Masry et~al.(2022)Masry, Long, Tan, Joty, and Hoque]{masry2022chartqa}
Ahmed Masry, Do~Xuan Long, Jia~Qing Tan, Shafiq Joty, and Enamul Hoque.
\newblock {ChartQA}: A benchmark for question answering about charts with
  visual and logical reasoning.
\newblock In \emph{Findings of the Association for Computational Linguistics:
  ACL 2022}, pages 2263--2279, 2022.
\newblock \doi{10.18653/v1/2022.findings-acl.177}.
\newblock URL \url{https://aclanthology.org/2022.findings-acl.177/}.

\bibitem[Masry et~al.(2025)Masry, Islam, Ahmed, Bajaj, Kabir, Kartha, Laskar,
  Rahman, Rahman, Shahmohammadi, Thakkar, Parvez, Hoque, and
  Joty]{masry2025chartqapro}
Ahmed Masry, Mohammed~Saidul Islam, Mahir Ahmed, Aayush Bajaj, Firoz Kabir,
  Aaryaman Kartha, Md~Tahmid~Rahman Laskar, Mizanur Rahman, Shadikur Rahman,
  Mehrad Shahmohammadi, Megh Thakkar, Md~Rizwan Parvez, Enamul Hoque, and
  Shafiq Joty.
\newblock {ChartQAPro}: A more diverse and challenging benchmark for chart
  question answering.
\newblock arXiv:2504.05506, 2025.
\newblock URL \url{https://arxiv.org/abs/2504.05506}.

\bibitem[Mukhopadhyay et~al.(2024)Mukhopadhyay, Qidwai, Garimella, Ramu, Gupta,
  and Roth]{mukhopadhyay2024unraveling}
Srija Mukhopadhyay, Adnan Qidwai, Aparna Garimella, Pritika Ramu, Vivek Gupta,
  and Dan Roth.
\newblock Unraveling the truth: Do {VLMs} really understand charts? a deep dive
  into consistency and robustness.
\newblock In \emph{Findings of the Association for Computational Linguistics:
  EMNLP 2024}, pages 16696--16717, 2024.
\newblock \doi{10.18653/v1/2024.findings-emnlp.973}.

\bibitem[{OpenAI}(2026)]{openai2026decisions}
{OpenAI}.
\newblock Decisions.
\newblock OpenAI API documentation, 2026.
\newblock URL \url{https://developers.openai.com/api/docs/guides/decisions}.
\newblock Accessed 24 October 2026.

\bibitem[Pai and Xian(2026)]{pai2026jevbench}
Dheeraj~Mohandas Pai and Lu~Xian.
\newblock {JevBench}: An open evaluation framework for typed decision models.
\newblock Preprints.org, version 2, not peer reviewed, 2026.

\bibitem[{Qwen Team}(2026)]{qwen2026qwen35}
{Qwen Team}.
\newblock {Qwen3.5-2B}: Model card.
\newblock Hugging Face, 2026.
\newblock URL \url{https://huggingface.co/Qwen/Qwen3.5-2B}.
\newblock Revision evaluated: 15852e8c16360a2fea060d615a32b45270f8a8fc;
  accessed 24 October 2026.

\bibitem[Ribeiro et~al.(2020)Ribeiro, Wu, Guestrin, and
  Singh]{ribeiro2020checklist}
Marco~Tulio Ribeiro, Tongshuang Wu, Carlos Guestrin, and Sameer Singh.
\newblock Beyond accuracy: Behavioral testing of {NLP} models with {CheckList}.
\newblock In \emph{Proceedings of the 58th Annual Meeting of the Association
  for Computational Linguistics}, pages 4902--4912, 2020.
\newblock \doi{10.18653/v1/2020.acl-main.442}.
\newblock URL \url{https://aclanthology.org/2020.acl-main.442/}.

\bibitem[Rodionov et~al.(2026{\natexlab{a}})Rodionov, Burguete-Lopez,
  Makarenko, Wang, Getman, and Fratalocchi]{rodionov2026moclip}
Sergei Rodionov, Arturo Burguete-Lopez, Maksim Makarenko, Qizhou Wang, Fedor
  Getman, and Andrea Fratalocchi.
\newblock {MOCLIP}: A foundation model for large-scale nanophotonic inverse
  design.
\newblock Figshare dataset, version 1, 2026{\natexlab{a}}.
\newblock URL
  \url{https://springernature.figshare.com/articles/dataset/MOCLIP_A_Foundation_Model_for_Large-Scale_Nanophotonic_Inverse_Design/30772553}.

\bibitem[Rodionov et~al.(2026{\natexlab{b}})Rodionov, Burguete-Lopez,
  Makarenko, Wang, Getman, and Fratalocchi]{rodionov2026moclippaper}
Sergei Rodionov, Arturo Burguete-Lopez, Maksim Makarenko, Qizhou Wang, Fedor
  Getman, and Andrea Fratalocchi.
\newblock {MOCLIP}: a foundation model for large-scale nanophotonic inverse
  design.
\newblock \emph{Nature Communications}, 17:\penalty0 9902, 2026{\natexlab{b}}.
\newblock \doi{10.1038/s41467-026-76714-x}.

\bibitem[Shen et~al.(2026)Shen, Lyu, Zhong, Li, Tao, Wang, Chen, and
  Fang]{shen2026specvqa}
Jialu Shen, Han Lyu, Suyang Zhong, Hanzheng Li, Haoyi Tao, Nan Wang, Changhong
  Chen, and Xi~Fang.
\newblock {SpecVQA}: A benchmark for spectral understanding and visual question
  answering in scientific images.
\newblock arXiv:2604.28039, 2026.
\newblock URL \url{https://arxiv.org/abs/2604.28039}.

\bibitem[Shi et~al.(2026)Shi, Liu, Zhao, Chi, Wang, Wang, and
  Feng]{shi2026trustdabench}
Boshen Shi, Yize Liu, Chen Zhao, Ce~Chi, Zhendong Wang, Xing Wang, and Junlan
  Feng.
\newblock {TrustDABench}: Benchmarking reliability and robustness of {LLMs} for
  structured data analysis.
\newblock arXiv:2608.24145, 2026.
\newblock URL \url{https://arxiv.org/abs/2608.24145}.

\bibitem[{vLLM Semantic Router}(2026)]{d3nano2026}
{vLLM Semantic Router}.
\newblock {d3-nano}: Model card.
\newblock Hugging Face, 2026.
\newblock URL \url{https://huggingface.co/vllm-sr/d3-nano}.
\newblock Checkpoint evaluated: 80820624a8170104679aa99897a6e9026eae0790; model
  card accessed 24 October 2026.

\bibitem[Wang et~al.(2025)Wang, Cao, Feng, Wang, and Cao]{wang2025gold}
Xiaobo Wang, Yu~Cao, Feng Feng, Ziyuan Wang, and Yuchen Cao.
\newblock Unified neural network model for predicting optical responses in gold
  nanostructures.
\newblock \emph{Journal of Applied Physics}, 137\penalty0 (6):\penalty0 063102,
  2025.
\newblock \doi{10.1063/5.0249206}.

\bibitem[Zheng et~al.(2024)Zheng, Zhou, Meng, Zhou, and
  Huang]{zheng2024selection}
Chujie Zheng, Hao Zhou, Fandong Meng, Jie Zhou, and Minlie Huang.
\newblock Large language models are not robust multiple choice selectors.
\newblock In \emph{International Conference on Learning Representations
  (ICLR)}, 2024.
\newblock URL \url{https://arxiv.org/abs/2309.03882}.
\newblock arXiv:2309.03882.

\bibitem[Zhou et~al.(2026)Zhou, Yang, and Zhao]{zhou2026pixeljev}
Xunlan Zhou, Xianliang Yang, and Li~Zhao.
\newblock From text decisions to pixels: An study of {Jev}-style visual choice
  model.
\newblock arXiv:2609.29283, 2026.
\newblock URL \url{https://arxiv.org/abs/2609.29283}.

\end{thebibliography}
